\documentclass[11pt]{article}

\usepackage{amsmath}
\usepackage{amssymb}
\usepackage[numbers]{natbib}
\usepackage[preprint]{acl}

\usepackage{times}
\usepackage{latexsym}
\usepackage{todonotes}
\usepackage[T1]{fontenc}

\usepackage[utf8]{inputenc}

\usepackage{microtype}

\usepackage{inconsolata}

\usepackage{graphicx}

\usepackage{makecell}

\usepackage[flushleft]{threeparttable}

\usepackage{multicol}

\title{LLMLagBench: Identifying Temporal Training Boundaries in Large Language Models}

\author{
 \textbf{Piotr Pęzik\textsuperscript{1}},
 \textbf{Konrad Kaczyński\textsuperscript{1}},
 \textbf{Maria Szymańska\textsuperscript{1}},
 \textbf{Filip Żarnecki\textsuperscript{1}},\\
 \textbf{Zuzanna Deckert\textsuperscript{1}},
 \textbf{Jakub Kwiatkowski\textsuperscript{1}},
 \textbf{Wojciech Janowski\textsuperscript{1}}
\\
 \textsuperscript{1} University of Lodz\\
\\
 \small{
   \textbf{Correspondence:} \href{mailto:piotr.pezik@uni.lodz.pl}{piotr.pezik@uni.lodz.pl}
 }
}

\begin{document}
\maketitle
\begin{abstract}

Large Language Models (LLMs) are pretrained on textual data up to a specific temporal cutoff. This creates a strict knowledge boundary beyond which models cannot provide accurate information without querying external sources. More subtly, when this limitation is unknown or ignored, LLMs may inadvertently blend outdated time-sensitive information with general knowledge during reasoning tasks, potentially compromising response accuracy. We introduce LLMLagBench, an LLM freshness benchmark, as a systematic approach for identifying the earliest probable temporal boundaries of an LLM's training data by evaluating its knowledge of recent events. We then apply this benchmark to evaluate a large set of LLMs, including models with both explicitly declared and undeclared training cutoffs. The reliability of the benchmark is assessed by manual validation and comparison with publicly released information about LLM pretraining.

\end{abstract}

\section{Introduction}

LLM parameters are learned from naturally-occurring textual data, instructions and preferences. All of these sources of data contain temporally-bounded knowledge about the world. Although LLMs can be continually pretrained on texts and further finetuned on instructions, their adaptation can be both impractical (especially in the case of large base models) and potentially risky due to the problem of catastrophic forgetting whereby incremental addition of new data impairs previously acquired knowledge or abilities of a pretrained model \cite{forget_Me_Not}.
As a result, major updates of base and instruct models are rare, and such models are effectively released with a strict knowledge cutoff date beyond which they are not updated until subsequent releases.\footnote{This problem is further explored in research on dynamically updating the knowledge base of LLMs \cite{li-goyal-2025-memorization} \cite{sun2025newdatapermeatesllm} \cite{zhao2025stylefactsmappingboundaries}.
}

The more obvious implications of this limitation, such as the tendency for LLMs to directly hallucinate about recent events, are usually addressed by providing retrieval-augmented answer generation mechanisms (RAG) \cite{ovadia_fine-tuning_2024}. However, there are potentially more subtle risks of interpolating outdated information with general knowledge in reasoning or forced-classification tasks. For example, a model can implicitly rely on recently outdated medical knowledge when making health-related recommendations. To illustrate, when Llama 3.1 70B-Instruct was indirectly prompted during our tests about whether blood lead levels of 15 µg/dL in manufacturing workers require intervention, it referenced outdated US OSHA standards without acknowledging that this value now represents the EU's biological limit threshold established in February 2024
occupational exposure regulations. The model provided guidance based on higher, less protective thresholds—all without indicating any uncertainty about its knowledge currency. This behavior illustrates how temporally-bounded knowledge can surreptitiously compromise decision-making in certain domains. Even models with access to search systems may either fail to recognize implicitly time-sensitive questions or retrieve incomplete results from search engines. It is therefore vital to know \textit{the earliest probable training cutoff} of any LLM\footnote{Our benchmark indicates October 2023 as a likely cutoff point for both LLama 3.1 and 3.3.} used to generate knowledge-based answers and decisions.

In practice, determining an LLM's actual knowledge cutoff is not straightforward. While a model's release date provides a hard upper bound for open-weight models, it is less reliable for models accessible only through vendor APIs. Even when developers declare a single specific cutoff date, interpreting its practical implications remains challenging. Furthermore, instruction-following models can exhibit contradictory behavior: they may declare overly conservative cutoff dates while simultaneously hallucinating about events far beyond their actual training period.\footnote{Table \ref{tab:model_cutoff_points} juxtaposes release dates, provider and model declarations.} The situation is further complicated by the fact that knowledge infusion may operate differently during pretraining, continued pretraining and post-training phases. Our analysis reveals that several LLMs exhibit multiple partial cutoff points, possibly corresponding to these distinct training stages. These observations motivate the need for a systematic empirical approach to identifying LLM knowledge boundaries as the central contribution of this work.

\section{Previous work}

A number of existing LLM evaluation benchmarks assess time-sensitive knowledge \cite{mousavi_dyknow_2024, holtermann2025world24hoursprobing}, temporal reasoning capibilities \cite{tan_towards_2023, tan_timelineqa_2023, chu_timebench_2024} and awareness \cite{zhu-etal-2025-evolvebench}.

Recent work by Zhu et al.~\cite{chenghaozhu-etal-2025-llm} introduced the concept of temporal generalization in LLMs, examining how models perform on data before and after their release dates. While their approach evaluates temporal bias and generalization by assuming the cutoff date equals the model release date (which is in fact the latest possible cutoff), the work reported in this paper   addresses a more fundamental question: empirically identifying when a model's training data actually ends. Our analysis reveals that training cutoffs often diverge significantly from release dates and that models frequently exhibit multiple partial cutoffs corresponding to different training phases.

Cheng et al.~\cite{cheng2024dateddatatracingknowledge} introduced perplexity-based probing to operationalize and identify ``effective cutoffs'' in LLMs, revealing significant temporal misalignment due to deduplication failures and CommonCrawl contamination with outdated content. Their approach requires white-box access to model probabilities, limiting evaluation to open-source models.

CutoffBench  (\url{https://cutoffbench.com}) is a publicly available benchmark which provides an automated approach to detecting LLM knowledge cutoffs using sports and financial time-series data with binary scoring and threshold detection. As explained further, in contrast to this benchmark we use PELT changepoint detection to enable identification of multiple partial cutoffs rather than assuming a single transition point. This is potentially important because many LLMs exhibit knowledge boundaries corresponding to different training phases. Second, our manually curated questions span diverse news domains rather than specialized areas like sports and finance, providing broader coverage of general knowledge.  We also explicitly track refusal behavior, distinguishing instruction-tuned refusal from genuine knowledge absence.

\section{\textit{LLMLagBench}}

This paper presents LLMLagBench, a benchmark designed to systematically identify the temporal boundaries of an LLM's training data through systematic evaluation of its knowledge of recent events. Using a set of manually selected questions densely sampled from news reports published over the last five years we calculate significant performance changepoints in answer accuracy.
\subsection{Data Collection and Filtering}
As of October 2025, the benchmark comprises 1,713 questions about events that could not be accurately answered before they were reported in news media. Representative examples of such question-answer pairs include:
\begin{itemize}
    \item (January 8, 2024) On what date did Adan Canto pass away? $\rightarrow$ Adan Canto passed away on January 8, 2024.
    \item (December 4, 2024) Who was fatally shot outside a New York City hotel early on December 4, 2024? $\rightarrow$ UnitedHealthcare CEO Brian Thompson was fatally shot outside a New York City hotel early on December 4, 2024.
    \item (April 3, 2024) What did John Legend perform with Ukrainian singer Mika Newton and poet Lyuba Yakimchuk during the 2022 Grammy Awards? $\rightarrow$ John Legend performed his song 'Free' with Ukrainian singer Mika Newton and poet Lyuba Yakimchuk during the 2022 Grammy Awards.
\end{itemize}
Although some of these questions could be guessed with some probability without access to later knowledge (including the one about the performance of Ukrainian artists), we considered the exact answers to these questions to be sufficiently unpredictable.

We sampled
a total of $\simeq$ 80,000 news items from the period 2021-2025 and clustered them by topic within each day. On average each selected story was reported
by 10 distinct news outlets (including at least one from our subset of worldwide/reliable sources). We then used \textbf{deepseek-ai/DeepSeek-V3-0324}
to extract approx. 8,400 candidate questions from these articles. Table \ref{tab:sources} summarizes the distribution of news outlets represented in the question set.

Despite detailed prompting, the LLM-based extractor occasionally generated questions that were trivial to answer, typically referring to regularly scheduled events or previously reported outcomes, as illustrated below:
\begin{itemize}
    \item (May 5, 2021) On what date in 2021 did Cinco de Mayo occur?
    \item (April 9, 2022) Which horse won the 2021 Grand National at Aintree?
\end{itemize}

To ensure benchmark quality, we manually validated all candidate questions and selected 1,713 question-answer pairs that met our criteria for temporal specificity and unpredictability. Table \ref{tab:questions_for_bench} presents additional examples illustrating the distinction between questions that meet our unpredictability criteria and those that were rejected as easily answerable.

\subsection{Answer Evaluation}
Each LLM evaluated in LLMLagBench is prompted to answer all 1,713 questions (approx. 7 questions per week) using the following standardized prompt:

\begin{quote}
\textit{Produce a concise answer to the following question. Generate only the answer, no additional comments. Don't speculate. If you don't know the answer, simply write 'I don't know'.}

\textit{Question: [QUESTION]}
\end{quote}

We then use \textbf{deepseek-ai/DeepSeek-V3-0324}
as an evaluator to assess these answers. The evaluation prompt instructs the LLM to rate each answer across three dimensions:

\begin{itemize}
    \item \textbf{Factual Accuracy} -- Are the facts in the answer correct? (0 = completely wrong or contains major factual errors; 1 = partially correct; 2 = fully accurate)
    \item \textbf{Relevance} -- Is all content related to the question? (0--2 scale with rating guidelines)
    \item \textbf{Faithfulness to the Gold Answer} -- Does the answer align with the reference answer? (0 = no overlap with the ideal answer; 1 = partial overlap, capturing some but not all core ideas; 2 = fully aligned with all core ideas)
\end{itemize}

Of these three criteria, LLMLagBench currently uses only \textbf{Faithfulness to the Gold Answer} as the primary evaluation metric. In a separate iteration, however, the evaluator model identifies instances where the evaluated models produce \textbf{refusals to answer} or otherwise evasive responses.

\begin{table}[h]
\centering
\begin{tabular}{lrr}
\hline
\textbf{Source} & \textbf{Candidates} & \textbf{Selected} \\
\hline
bbc.com & 325 & 58 \\
cnn.com & 200 & 43 \\
dailymail.co.uk & 1529 & 277 \\
foxnews.com & 151 & 41 \\
independent.co.uk & 2862 & 594 \\
nypost.com & 839 & 169 \\
nytimes.com & 454 & 81 \\
theguardian.com & 783 & 187 \\
usatoday.com & 456 & 98 \\
washingtonpost.com & 301 & 62 \\
washingtontimes.com & 515 & 103 \\
\hline
\textbf{Total} & \textbf{8,415} & \textbf{1,713} \\
\hline
\end{tabular}
\caption{Question sources and selection statistics.}
\label{tab:sources}
\end{table}

To assess the reliability of the automated evaluation, we measured inter-rater agreement between the evaluator model’s faithfulness ratings and those provided by two human annotators on a subset of 500 questions across different versions of the evaluation prompts. The final configuration achieved Cohen’s Kappa scores of 0.81 and 0.83, respectively.
The generation temperature of the evaluator model was tuned to maximize these inter-rater agreement values.

\subsection{Training Cutoff Detection}
Using the optimized generation hyperparameters of the evaluator model, we calculated faithfulness scores ranging from 0 to 2 for all evaluated models. While these scores can be averaged over specific time periods to estimate a model's performance on time-sensitive knowledge questions, the primary objective of LLMLagBench is to provide an independent estimate of the earliest probable training cutoff point.

Our approach is based on the assumption that models should exhibit a sudden, significant decrease in performance beyond their training data boundary. To identify such discontinuities, we employ the Pruned Exact Linear Time (PELT) changepoint detection algorithm \cite{Killick_2012}, which efficiently detects multiple changepoints in time series data by minimizing a penalized cost function. PELT has been widely used for identifying structural breaks in sequential data \cite{aminikhanghahi2017survey} and is particularly well-suited for our task as it can detect abrupt shifts in mean performance without requiring prior specification of the number of changepoints. We apply PELT to the time-ordered sequence of faithfulness scores, allowing the algorithm to identify dates where model performance exhibits statistically significant drops.

\begin{figure*}[!htbp]
    \centering
    \includegraphics[width=\textwidth]{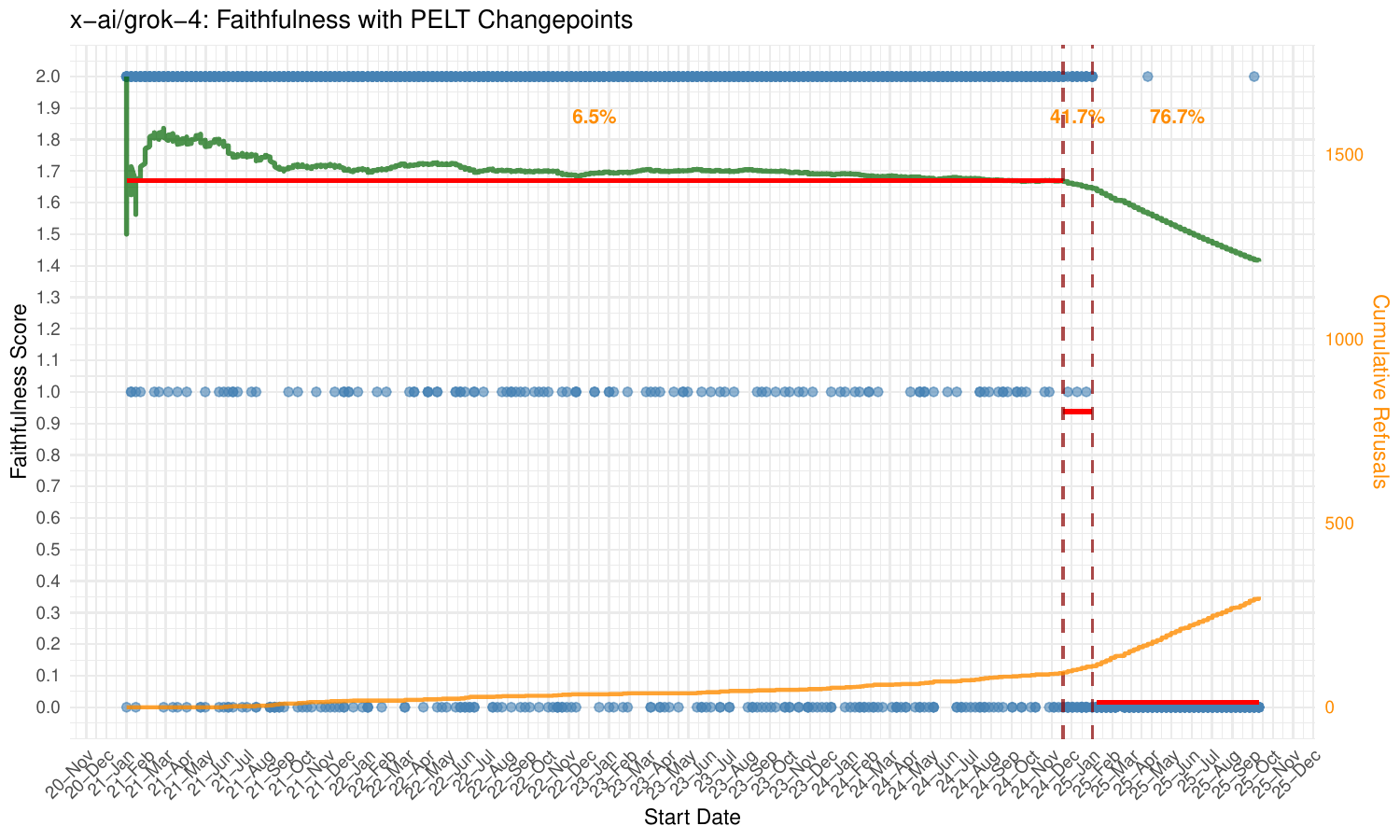}
    \caption{xAI Grok-4 training cutoff point estimation.}
    \label{fig:x-ai-grok-4_pelt_with_refusals}
\end{figure*}

It should be additionally noted that detecting the true pretraining cutoff is complicated by the fact that instruction-tuned models may refuse to answer questions beyond a declared cutoff date, despite having been pretrained on data from later time periods. To address this caveat, we also calculate average and cumulative refusal rates alongside faithfulness scores.

\section{Results and Findings}
We applied LLMLagBench to a diverse set of LLMs to evaluate how well our methodology identifies training cutoffs across different scenarios. Our analysis reveals several distinct patterns in the relationship between model release dates, vendor- and mode-declared cutoffs, and empirically detected training boundaries. In the following subsections, we present selected cases that illustrate these patterns, demonstrating how LLMLagBench can uncover information about training data freshness that may not be evident from public disclosures or direct model queries. Each case highlights a different combination of available information (release dates, declared cutoffs, self-reported dates) and our benchmark-derived estimates.

\subsection{Grok-4: Single Cutoff Detection}
Figure \ref{fig:x-ai-grok-4_pelt_with_refusals} illustrates the combined effect of this methodology when applied to xAI Grok-4. Blue dots represent faithfulness scores (0-2 scale, left y-axis) for individual questions ordered by event date, while red horizontal lines indicate mean faithfulness scores within segments identified by PELT. The green curve shows the cumulative average faithfulness score at each week. Red dashed vertical lines mark the detected changepoints (possible training boundaries), and the orange line (right y-axis) tracks cumulative refusals over time. The model maintains high performance (faithfulness $\approx$ 1.7) from the beginning 2021 through late 2024, confirming comprehensive knowledge of events during this period. However, PELT identifies two closely localized changepoints in late 2024 and early 2025, where performance degrades progressively as refusal rates increase from 6.5\% to 77.0\%. This refusal pattern is characteristic of instruction-tuned models trained to decline questions about events beyond a declared cutoff date. The sharp rise in refusals coinciding with the drop in faithfulness scores suggests that Grok-4's behavioral cutoff aligns closely with its actual pretraining data boundary. Given the overall performance characteristics, \textit{the earliest probable cutoff date} for Grok-4 can be estimated at late 2024. As shown in Table \ref{tab:model_cutoff_points}, this estimate revealed by LLMLagBench is significantly earlier than the model's release date (August 2025) and it coincides with the xAI's declarations of November 2024 as the cutoff for this model. Interestingly, this changepoint cannot be reliably determined by directly prompting the model, as in our tests Grok-4 self-reports a training cutoff of November 2023---more than a year earlier than our empirical estimate suggests.

\subsection{Claude Sonnet 4: Multiple Partial Cutoffs}

\begin{figure*}[!htbp]
    \centering
    \includegraphics[width=\textwidth]{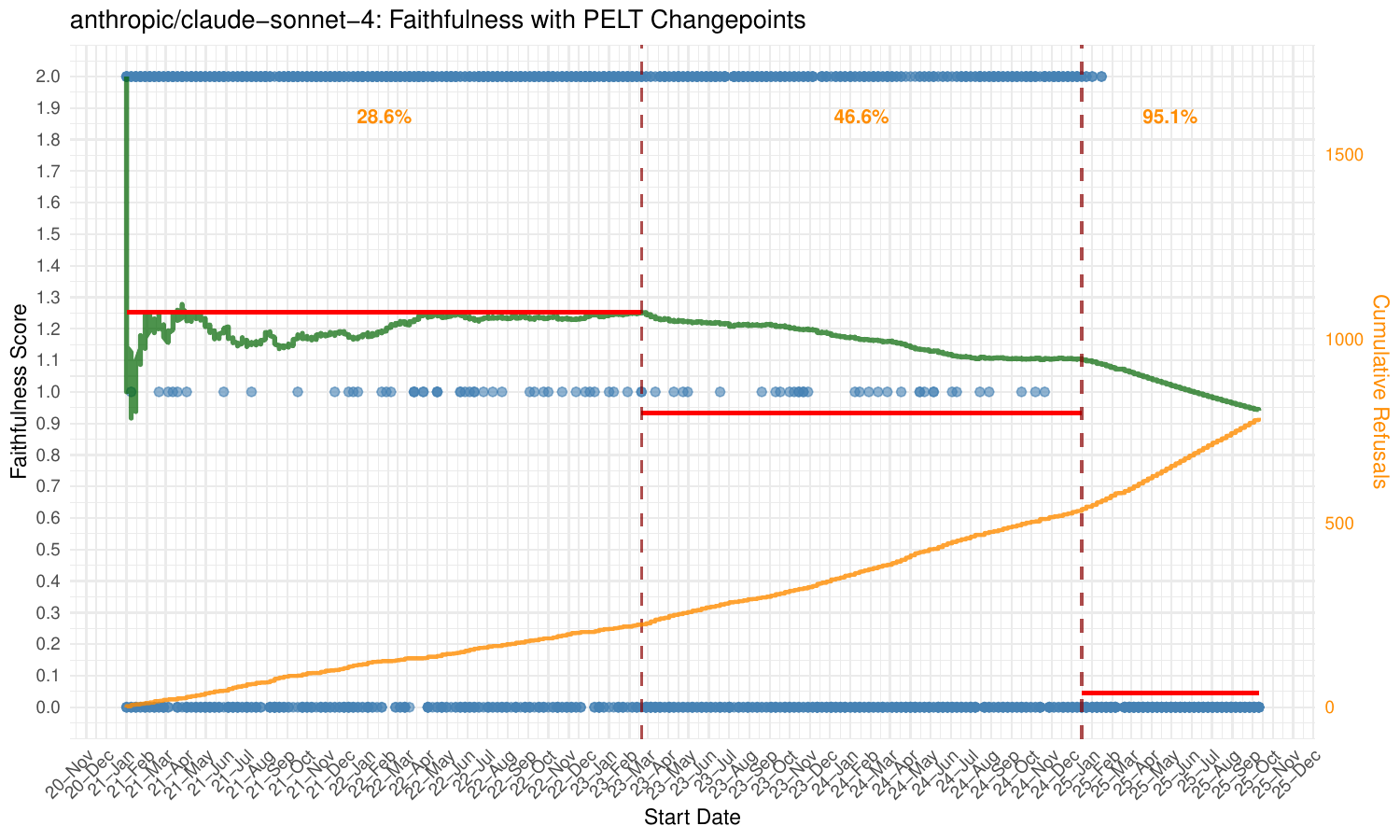}
    \caption{Claude Sonnet 4 training cutoff point estimation.}
    \label{fig:anthropic-claude-sonnet-4_pelt_with_refusals}
\end{figure*}

Figure \ref{fig:anthropic-claude-sonnet-4_pelt_with_refusals} presents a different pattern for Anthropic's Claude Sonnet 4. PELT identifies two distinct changepoints at February 2023 and December 2024, dividing the model's performance into three segments with progressively declining mean faithfulness scores (1.25, 0.93, and 0.05, respectively) and increasing refusal rates (28.7\%, 46.8\%, and 95.1\%).

The first changepoint in early 2023 is particularly notable, occurring well before the model provider's declared cutoff of January 2025 and its release date of May 2025. This suggests that Claude Sonnet 4's knowledge base was not uniformly updated throughout its training process. The extended intermediate segment (February 2023 to December 2024) exhibits moderate performance with nearly half of all questions refused, indicating partial knowledge coverage during this period. This pattern may indicate that the model in question underwent initial pretraining with a cutoff in early 2023, followed by selective knowledge updates or continued pretraining on limited data extending into late 2024.

The second changepoint at December 2024 aligns closely with the provider-declared cutoff of January 2025, where performance drops sharply and refusal rates exceed 95\%. Unlike Grok-4, which maintained consistently high performance until its cutoff, Claude Sonnet 4 exhibits multiple partial knowledge boundaries possibly corresponding to different training phases. If this interpretation is correct, the declared January 2025 cutoff may obscure the significant knowledge degradation that began nearly two years earlier.

\subsection{GPT-OSS-120B: Discrepancy Between Declared and Detected Cutoffs}

\begin{figure*}[!htbp]
    \centering
    \includegraphics[width=\textwidth]{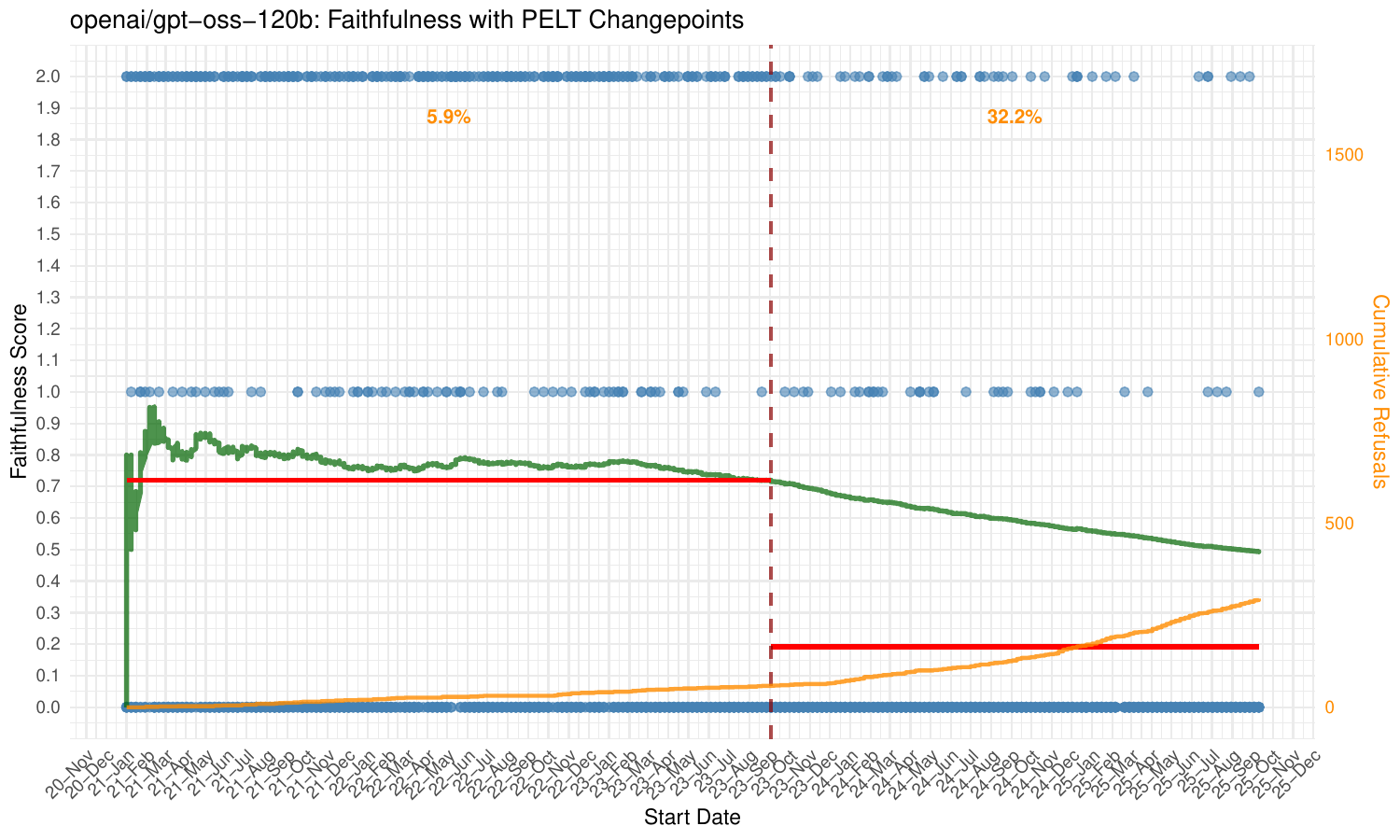}
    \caption{GPT-OSS-120B training cutoff point estimation.}
    \label{fig:openai-gpt-oss-120b_pelt_with_refusals}
\end{figure*}

Figure \ref{fig:openai-gpt-oss-120b_pelt_with_refusals} illustrates yet another distinct pattern in training cutoff characteristics. PELT identifies a single changepoint at September 2023, dividing performance into two segments with mean faithfulness scores of 0.72 and 0.19, and refusal rates of 5.9\% and 32.2\%, respectively.

The benchmark-detected cutoff of September 2023 reveals a significant discrepancy with OpenAI's declared cutoff of July 2024—nearly a full year later. This suggests that while the provider may have included some training data extending into mid-2024, the model's reliable knowledge effectively ends in September 2023. The relatively modest initial faithfulness score of 0.72 (compared to Grok-4's 1.7 or Claude Sonnet 4's 1.25) indicates that even within its primary knowledge period, GPT-OSS-120B demonstrates lower overall performance on time-sensitive factual questions.

Notably, the model's self-reported cutoff of September 2021 is dramatically earlier than both the provider's declaration and our empirical estimate—approximately two years off from the benchmark-detected boundary. This extreme underestimation suggests either highly conservative instruction-tuning regarding knowledge claims, or a misalignment between the model's self-assessment capabilities and its actual training data scope.

The post-cutoff segment shows a moderate increase in refusals (from 5.9\% to 32.2\%), but the refusal rate remains substantially lower than those observed in Grok-4 or Claude Sonnet 4. This indicates that GPT-OSS-120B continues to attempt answers for post-cutoff questions rather than reliably declining them, potentially leading to higher rates of hallucination about recent events. This case underscores the risk of relying on either provider declarations or model self-reports without empirical validation through benchmarks like LLMLagBench.
\subsection{Gemma 3 Models: Model Size and Cutoff Detection Reliability}

\begin{figure*}[!htbp]
    \centering
    \includegraphics[width=\textwidth]{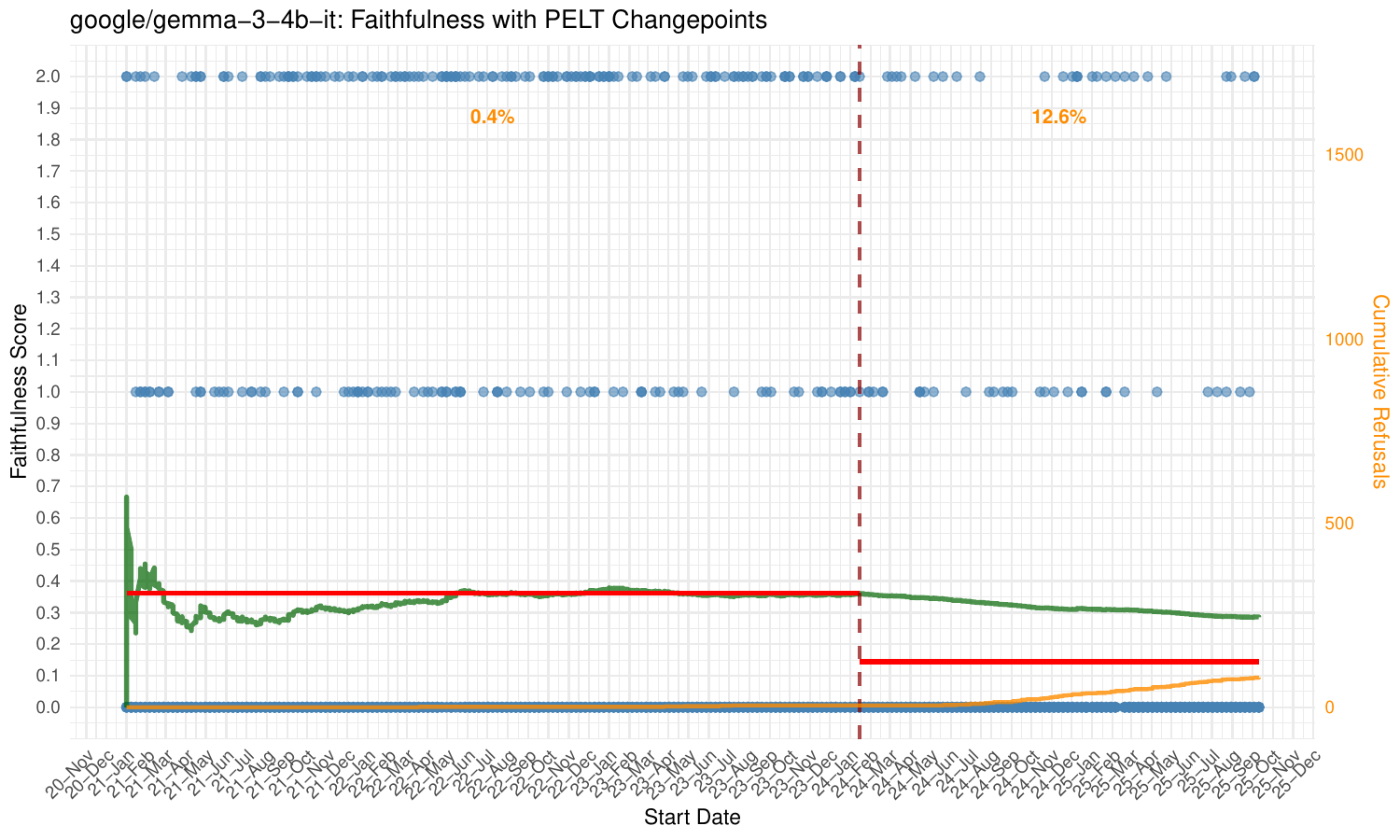}
    \caption{Gemma-3-4B training cutoff point estimation.}
    \label{fig:google-gemma-3-4b-it_pelt_with_refusals}
\end{figure*}

\begin{figure*}[!htbp]
    \centering
    \includegraphics[width=\textwidth]{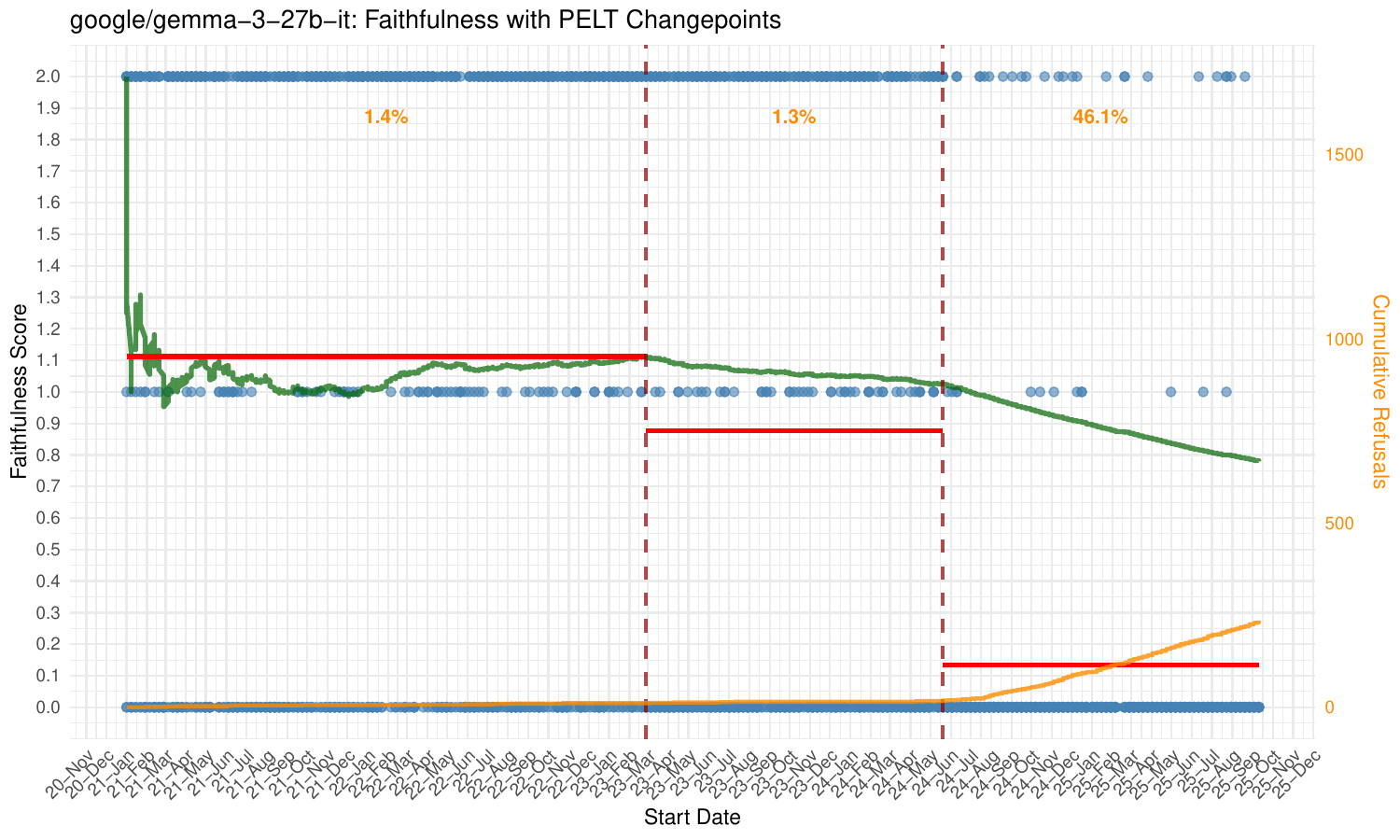}
    \caption{Gemma-3-27B training cutoff point estimation.}
    \label{fig:google-gemma-3-27b-it_pelt_with_refusals}
\end{figure*}

Figures \ref{fig:google-gemma-3-4b-it_pelt_with_refusals} and \ref{fig:google-gemma-3-27b-it_pelt_with_refusals} present results for Google's Gemma 3 models in two different sizes (4B and 27B parameters), offering insights into how model capacity affects both overall performance and the reliability of cutoff detection.

The larger Gemma 3-27B model demonstrates a pattern similar to Claude Sonnet 4, with two changepoints at February 2023 and May 2024. The model exhibits stronger initial performance (faithfulness score of 1.11), followed by moderate performance in the intermediate segment (0.88), and finally sharp degradation after May 2024 (0.13 with 46.1\% refusals). Notably, the benchmark-detected cutoff of May 2024 aligns closely with Google's declared cutoff of August 2024—a discrepancy of only three months.

For the smaller Gemma 3-4B model PELT identifies a single changepoint at January 2024, with substantially lower faithfulness scores both before (0.36) and after (0.14) the cutoff. However, despite this low absolute performance, the changepoint remains clearly detectable, and refusal rates increase from near-zero (0.4\%) to 12.6\% after the cutoff. Notably, the post-cutoff refusal rate of 12.6\% remains substantially lower than that of the larger 27B variant (46.1\%), suggesting that smaller models may be more prone to hallucination about post-cutoff events. More generally, model size significantly impacts absolute performance on time-sensitive factual questions as the 4B model struggles even within its training data range.

\begin{table*}
  \centering
  \resizebox{\textwidth}{!}{\begin{tabular}{lccccccc}
    \hline
    \textbf{Model name} & \textbf{Changepoints} & \textbf{Avg faithfulness} & \textbf{Refusal rate} & \textbf{1st changepoint} & \textbf{1st segment avg} & \textbf{2nd changepoint} & \textbf{Post-changepoint refusal} \\
    \hline
    x-ai/grok-4 & 2 & 1.41 & 0.18 & 2024-11-18 & 1.67 & 2025-01-01 & 0.77 \\
    x-ai/grok-3 & 1 & 1.37 & 0.10 & 2024-12-16 & 1.61 & -- & 0.58 \\
    moonshotai/kimi-k2 & 2 & 1.25 & 0.15 & 2023-02-26 & 1.56 & 2025-01-22 & 0.67 \\
    deepseek-ai/DeepSeek-R1-0528 & 1 & 1.14 & 0.18 & 2024-07-15 & 1.48 & -- & 0.61 \\
    google/gemini-2.0-flash-001 & 1 & 1.10 & 0.27 & 2024-06-24 & 1.47 & -- & 0.86 \\
    anthropic/claude-3-5-sonnet-20240620 & 2 & 1.07 & 0.35 & 2023-09-03 & 1.62 & 2024-03-11 & 0.87 \\
    z-ai/glm-4.5 & 2 & 1.02 & 0.23 & 2024-03-11 & 1.29 & 2025-01-01 & 0.72 \\
    anthropic/claude-sonnet-4 & 2 & 0.94 & 0.46 & 2023-02-19 & 1.25 & 2024-12-16 & 0.95 \\
    mistralai/mistral-medium-3.1 & 2 & 0.91 & 0.29 & 2023-02-26 & 1.35 & 2024-07-15 & 0.83 \\
    google/gemini-2.5-flash & 2 & 0.90 & 0.46 & 2024-03-25 & 1.25 & 2024-06-17 & 0.99 \\
    deepseek-ai/DeepSeek-V3-0324 & 2 & 0.89 & 0.40 & 2023-09-17 & 1.27 & 2024-07-01 & 0.93 \\
    deepseek-ai/DeepSeek-Chat-V3.1 & 1 & 0.89 & 0.31 & 2024-07-22 & 1.15 & -- & 0.85 \\
    openai/gpt-4o-2024-08-06 & 1 & 0.88 & 0.50 & 2023-10-29 & 1.46 & -- & 0.99 \\
    mistralai/mistral-medium-3 & 2 & 0.88 & 0.37 & 2023-02-26 & 1.32 & 2024-07-15 & 0.91 \\
    anthropic/claude-3-opus-20240229 & 2 & 0.80 & 0.52 & 2023-04-02 & 1.52 & 2023-08-27 & 0.96 \\
    google/gemma-3-27b-it & 2 & 0.78 & 0.14 & 2023-02-26 & 1.11 & 2024-05-20 & 0.46 \\
    meta-llama/Meta-Llama-3.1-70B-Instruct & 2 & 0.74 & 0.43 & 2023-02-19 & 1.44 & 2023-09-03 & 0.86 \\
    meta-llama/Llama-3.3-70B-Instruct & 2 & 0.70 & 0.52 & 2023-02-26 & 1.41 & 2023-09-03 & 0.96 \\
    amazon/nova-pro-v1 & 2 & 0.69 & 0.42 & 2023-02-26 & 1.11 & 2024-05-20 & 0.91 \\
    ai21/jamba-large-1.7 & 2 & 0.64 & 0.61 & 2023-02-19 & 1.04 & 2024-02-12 & 0.95 \\
    mistralai/Mixtral-8x22B-Instruct-v0.1 & 3 & 0.62 & 0.28 & 2022-08-13 & 1.36 & 2023-02-12 & 0.39 \\
    openai/gpt-4o-mini-2024-07-18 & 2 & 0.60 & 0.57 & 2023-02-26 & 1.09 & 2023-09-24 & 0.96 \\
    CohereForAI/c4ai-command-a-03-2025 & 2 & 0.59 & 0.62 & 2023-02-19 & 1.04 & 2024-04-15 & 0.97 \\
    CohereForAI/c4ai-command-r-plus & 2 & 0.58 & 0.39 & 2022-05-21 & 1.31 & 2023-01-15 & 0.61 \\
    openai/gpt-oss-120b & 1 & 0.49 & 0.17 & 2023-09-03 & 0.72 & -- & 0.32 \\
    meta-llama/Llama-4-Scout-17B-16E-Instruct & 2 & 0.48 & 0.42 & 2023-01-29 & 0.78 & 2024-03-11 & 0.83 \\
    gpt-3.5-turbo-0125 & 2 & 0.44 & 0.42 & 2021-09-17 & 1.40 & 2022-04-09 & 0.54 \\
    anthropic/claude-3-haiku-2024030 & 1 & 0.40 & 0.74 & 2023-01-29 & 0.86 & -- & 0.97 \\
    google/gemma-3-4b-it & 1 & 0.29 & 0.05 & 2024-01-15 & 0.36 & -- & 0.13 \\
    meta-llama/Llama-3.1-8B-Instruct & 1 & 0.28 & 0.55 & 2023-03-26 & 0.56 & -- & 0.81 \\
    openai/gpt-oss-20b & 1 & 0.24 & 0.15 & 2023-07-30 & 0.34 & -- & 0.28 \\
    openchat/openchat-3.5-1210 & 2 & 0.20 & 0.65 & 2021-04-30 & 0.81 & 2023-01-08 & 0.83 \\
    amazon/nova-lite-v1 & 1 & 0.18 & 0.72 & 2023-02-12 & 0.35 & -- & 0.86 \\
    mistralai/ministral-8b & 1 & 0.13 & 0.80 & 2023-01-22 & 0.27 & -- & 0.92 \\
    Qwen/Qwen2.5-Omni-7B & 0 & 0.04 & 0.87 & -- & -- & -- & -- \\
    \hline
  \end{tabular}}
  \caption{Summary of model faithfulness analysis with changepoint detection.}
  \label{tab:model_summary}
\end{table*}

\begin{table*}
  \centering
  \resizebox{\textwidth}{!}{
  \begin{threeparttable}
  \begin{tabular}{llllll}
    \hline
    \textbf{Model name} & \textbf{Parameters} & \textbf{Release date} & \textbf{Provider cutoff} & \textbf{Model cutoff} & \textbf{LLMLagBench} \\
    \hline
    ai21/jamba-large-1.7 & 398B & 2024.07.03 \tnote{1} & \makecell[l]{2024.03 \tnote{1} * \\ 2024.08.22 \tnote{2} *} & 2024.02 & 2023.02/2024.02\\
    amazon/nova-lite-v1 & N/A & 2025.03.17 \tnote{3} & N/A \tnote{3} & 2021.10 & 2023.02\\
    amazon/nova-pro-v1 & N/A & 2025.03.17 \tnote{3} & N/A \tnote{3} &  2023.10 & 2023.02/2024.05\\
    anthropic/claude-3-5-sonnet-20240620 & N/A & 2024.08.21 \tnote{4} & 2024.04 \tnote{4} & 2022.09 & 2023.09/2024.03\\
    anthropic/claude-3-haiku-2024030 & N/A & 2024.03.13 \tnote{5} & 2023.08 \tnote{6} & 2021 & 2023.01\\
    anthropic/claude-3-opus-20240229 & N/A & 2024.03.04 \tnote{7} & 2023.08 \tnote{6} & 2021 & 2023.04/2023.08\\
    anthropic/claude-sonnet-4 & N/A & 2025.05.22 \tnote{8} & 2025.01 \tnote{9} & 2024.04 & 2023.02/2024.12\\
    CohereForAI/c4ai-command-a-03-2025 & 111B & 2025.03.13 \tnote{10} & 2024.07.01 \tnote{11} & 2024.06 & 2023.02/2024.04\\
    CohereForAI/c4ai-command-r-plus & 35B & 2024.08 \tnote{12} & 2024.07.01 \tnote{13} & 2023.01 & 2022.05/2023.02\\
    deepseek-ai/DeepSeek-V3-0324 & 685B & 2024.12.27 \tnote{14} & N/A \tnote{14} & 2023.09 & 2024.07\\
    deepseek-ai/DeepSeek-Chat-V3.1 & 671B & 2025.08.21 \tnote{15} & N/A \tnote{15} & 2023.10 & 2024.07\\
    deepseek-ai/DeepSeek-R1-0528 & 685B & 2025.01.22 \tnote{16} & N/A \tnote{16} & 2024.07 & 2024.07\\
    google/gemma-3-27b-it & 27B & 2025.08.14 \tnote{17} & 2024.08 \tnote{18} & 2023.11.09 & 2023.02/2024.05\\
    google/gemma-3-4b-it & 4B & 2025.08.14 \tnote{17} & 2024.08 \tnote{18} & 2023.11.09 & 2024.01\\
    google/gemini-2.0-flash-001 & N/A & 2024.12.11 \tnote{19} & 2024.06 \tnote{20} & 2021.09/2023 & 2024.06\\
    google/gemini-2.5-flash & N/A & 2025.07.22 \tnote{20} & 2025.01 \tnote{20} & 2024.07 & 2024.03/2024.06\\
    openai/gpt-3.5-turbo-0125 & 175B & 2022.11.30 \tnote{21} & 2021.09.01 \tnote{22} & 2021.09 & 2021.09/2022.04\\
    openai/gpt-4o-2024-08-06 & N/A & 2024.08.06 \tnote{23} & 2023.10.01 \tnote{24} & 2023.10 & 2023.10\\
    openai/gpt-4o-mini-2024-07-18 & N/A & 2024.07.18 \tnote{25} & 2023.10.01 \tnote{26} & 2021.10 & 2023.02/2023.09\\
    meta-llama/Llama-3.1-8B-Instruct & 8B & 2024.07.21 \tnote{27} & 2023.12 \tnote{28} & 2023.12 & 2023.03\\
    meta-llama/Llama-3.3-70B-Instruct & 70B & 2024.12.06 \tnote{29} & 2023.12 \tnote{29} & 2023.12 & 2023.02/2023.09\\
    meta-llama/Llama-4-Scout-17B-16E-Instruct & 17B & 2025.04.05 \tnote{30} & 2024.08 \tnote{31} & 2024.08 & 2023.01/2024.03\\
    meta-llama/Meta-Llama-3.1-70B-Instruct & 70B & 2024.07.21 \tnote{27} & 2023.12 \tnote{28} & 2023.12 & 2023.02/2023.09\\
    mistralai/ministral-8b & 8B & 2024.10.09 \tnote{32} & N/A \tnote{32} & 2023.10 & 2023.01\\
    mistralai/mistral-medium-3 & N/A & 2025.05.07 \tnote{33} & N/A \tnote{33} & 2023.12 & 2023.02/2024.07\\
    mistralai/mistral-medium-3.1 & N/A & 2025.08.12 \tnote{34} & N/A \tnote{34} & 2024.11 & 2023.02/2024.07\\
    mistralai/Mixtral-8x22B-Instruct-v0.1 & 22B & 2024.01.08 \tnote{35} & N/A \tnote{35} & 2023 & 2022.08/2023.02/2023.06\\
    moonshotai/kimi-k2 & 1T & 2025.08.28 \tnote{36} & 2025.07 \tnote{37} & 2025.04 & 2023.02/2025.01\\
    openai/gpt-oss-120b & 120B & 2025.08.05 \tnote{38} & 2024.07 \tnote{38} & 2021.09 & 2023.09\\
    openai/gpt-oss-20b & 20B & 2025.08.05 \tnote{38} & 2024.07 \tnote{38} & 2021.09 & 2023.07\\
    openchat/openchat-3.5-1210 & 7B & 2024.01.06 \tnote{39} & N/A \tnote{40} & 2021.09 & 2021.04/2023.01\\
    x-ai/grok-3 & N/A & 2025.02.19 \tnote{41} & 2024.11 \tnote{42} & 2023.10 & 2024.12\\
    x-ai/grok-4 & N/A & 2025.08.20 \tnote{43} & 2024.11 \tnote{42} & 2023.10 & 2024.11/2025.01\\
    z-ai/glm-4.5 & 355B & 2025.08.08 \tnote{44} & N/A \tnote{44} & 2024.06 & 2024.03/2025.01\\
    Qwen/Qwen2.5-Omni-7B & 7B & 2025.01.03 \tnote{45} & N/A \tnote{45} & 2023.12 & --\\
    \hline
  \end{tabular}
    \caption{
    Model cutoff dates according to developers, the model itself, and LLMLagBench predictions. In entries marked with *, the provider specifies more than one date, depending on the source used. If there is more than one date present under LLMLagBench prediction, multiple PELT changepoints have been detected.
  }
  \label{tab:model_cutoff_points}
\begin{tablenotes}
\scriptsize
  \begin{tabular}{p{0.142\textwidth} p{0.142\textwidth} p{0.142\textwidth} p{0.142\textwidth} p{0.142\textwidth} p{0.142\textwidth} p{0.142\textwidth}}
    \item[1] \hspace{0.1em} \citep{lieber_jamba_2024} &
    \item[2] \hspace{0.1em} \citep{jamba-hf} &
    \item[3] \hspace{0.1em} \citep{agi_amazon_2025} &
    \item[4] \hspace{0.1em} \citep{claude-3.5} &
    \item[5] \hspace{0.1em} \citep{claude-3-haiku-release} &
    \item[6] \hspace{0.1em} \citep{claude-3} &
    \item[7] \hspace{0.1em} \citep{claude-3-opus-release} \\

    \item[8] \hspace{0.1em} \citep{claude-4-release} &
    \item[9] \hspace{0.1em} \citep{claude-4} &
    \item[10] \hspace{0.2em} \citep{command-a-release} &
    \item[11] \hspace{0.2em} \citep{cohere_command_2025} &
    \item[12] \hspace{0.2em} \citep{command-r} &
    \item[13] \hspace{0.1em} \citep{cohere_command_r} &
    \item[14] \hspace{0.05em} \citep{deepseek-ai_deepseek-v3_2025} \\

    \item[15] \hspace{0.2em} \citep{deepseek-3.1} &
    \item[16] \hspace{0.05em} \citep{deepseek-ai_deepseek-r1_2025} &
    \item[17] \hspace{0.2em} \citep{gemma-release} &
    \item[18] \hspace{0.2em} \citep{gemma-3} &
    \item[19] \hspace{0.2em} \citep{gemini-2-release} &
    \item[20] \hspace{0.2em} \citep{comanici_gemini_2025} &
    \item[21] \hspace{0.2em} \citep{gpt-3.5-release} \\

    \item[22] \hspace{0.2em} \citep{gpt-3.5} &
    \item[23] \hspace{0.2em} \citep{gpt-4o-release} &
    \item[24] \hspace{0.2em} \citep{gpt-4o} &
    \item[25] \hspace{0.2em} \citep{gpt-4o-mini-release} &
    \item[26] \hspace{0.2em} \citep{gpt-4o-mini} &
    \item[27] \hspace{0.2em} \citep{llama-3.1-release} &
    \item[28] \hspace{0.2em} \citep{llama-3.1} \\

    \item[29] \hspace{0.2em} \citep{llama-3.3} &
    \item[30] \hspace{0.2em} \citep{llama-4-release} &
    \item[31] \hspace{0.2em} \citep{llama-4-scout} &
    \item[32] \hspace{0.2em} \citep{ministral-release} &
    \item[33] \hspace{0.2em} \citep{mistral} &
    \item[34] \hspace{0.2em} \citep{mistral-medium-3.1-release} &
    \item[35] \hspace{0.2em} \citep{jiang_mixtral_2024} \\

    \item[36] \hspace{0.2em} \citep{kimi-release} &
    \item[37] \hspace{0.2em} \citep{kimi-sim} &
    \item[38] \hspace{0.2em} \citep{openai_gpt-oss-120b_2025} &
    \item[39] \hspace{0.2em} \citep{openchat-release} &
    \item[40] \hspace{0.2em} \citep{openchat} &
    \item[41] \hspace{0.2em} \citep{grok3} &
    \item[42] \hspace{0.2em} \citep{grok} \\

    \item[43] \hspace{0.2em} \citep{grok4} &
    \item[44] \hspace{0.2em} \citep{team_glm-45_2025} &
    \item[45] \hspace{0.2em} \citep{qwen-2.5} \\
  \end{tabular}
\end{tablenotes}

  \end{threeparttable}
  }
\end{table*}

\begin{table*}
  \centering
  \resizebox{\textwidth}{!}{
  \begin{tabular}{lllll}
    \hline
    \textbf{\#} & \textbf{Date} & \textbf{Question} & \textbf{Gold Answer} & \textbf{Possible decision}\\
    \hline
    1 & 2022.01.01 & \makecell[l]{What was Sir Arthur Conan Doyle doing \\ according to the 1921 census in the UK?} & \makecell[l]{Sir Arthur Conan Doyle appeared \\ to be holding a séance.} & Accepted \\
    2 & 2022.02.26 & \makecell[l]{What role did Volodymyr Zelensky voice \\ in the Ukrainian release of the *Paddington* film \\ before he became Ukraine's president?} & \makecell[l]{Volodymyr Zelensky was the voice of \\ Paddington Bear in the Ukrainian release \\ of the *Paddington* film.} & Rejected \\
    3 & 2022.02.05 & \makecell[l]{On February 8, 2022, what was Eileen Gu's \\ total score in the women's Big Air competition \\ at the Beijing Winter Olympics?} & \makecell[l]{Eileen Gu's total score in the women's \\ Big Air competition at the Beijing Winter Olympics \\ on February 8, 2022, was 188.25.} & Accepted \\
    4 & 2022.04.09 & \makecell[l]{Which horse won the 2021 Grand National \\ at Aintree?} & \makecell[l]{Minella Times, ridden by Rachael Blackmore, \\ won the 2021 Grand National at Aintree.} & Rejected \\
    5 & 2022.04.02 & \makecell[l]{What did John Legend perform with \\ Ukrainian singer Mika Newton and poet \\ Lyuba Yakimchuk during the 2022 Grammy Awards?} & \makecell[l]{John Legend performed his song *Free* \\ with Ukrainian singer Mika Newton and poet \\ Lyuba Yakimchuk during the 2022 Grammy Awards.} & Accepted \\
    6 & 2022.07.23 & \makecell[l]{What happened to the Sky News interview \\ with Nadine Dorries on July 28, 2022, \\ in Birmingham?} & \makecell[l]{The Sky News interview with Nadine Dorries \\ was forced off the air due to an incident behind \\ the camera involving a man shouting \\ at the camera operator.} & Accepted \\
    7 & 2022.08.27 & \makecell[l]{What was the US life expectancy in 2021, \\ and how much did it decline compared to 2020?} & \makecell[l]{In 2021, the US life expectancy was about \\ 76 years and 1 month, declining by nearly a year \\ from 2020 when it was 77 years.} & Rejected \\
    8 & 2022.09.17 & \makecell[l]{On what date did Queen Elizabeth II's \\ state funeral take place?} & September 19, 2022 & Rejected \\
    9 & 2022.09.24 & \makecell[l]{On what date in 2022 did Rosh Hashanah, \\ the Jewish New Year, begin?} & September 25, 2022 & Rejected \\
    10 & 2022.11.19 & \makecell[l]{On what date was Bob Iger announced \\ as the returning CEO of Walt Disney Co.?} & November 20, 2022 & Accepted \\
    11 & 2022.12.10 & \makecell[l]{On what date did the world premiere of \\ *Avatar: The Way of Water* take place \\ at the Odeon Leicester Square?} & \makecell[l]{The world premiere of *Avatar: The Way of Water* \\ took place at the Odeon Leicester Square \\ on December 6, 2022.} & Rejected \\
    12 & 2023.01.15 & \makecell[l]{On January 17, 2023, what was the final score \\ of the Australian Open men's singles match \\ between Andy Murray and Matteo Berrettini?} & \makecell[l]{Andy Murray defeated Matteo Berrettini \\ 6-3 6-3 4-6 6-7(7-9) 7-6(10-6) \\ in the opening round of the 2023 Australian Open.} & Accepted \\
    13 & 2024.01.08 & \makecell[l]{On what date did Adan Canto pass away?} & \makecell[l]{Adan Canto passed away on \\ January 8, 2024.} & Accepted \\
    14 & 2024.02.19 & \makecell[l]{Which two pandas from the San Diego Zoo \\ were returned to China in 2019?} & \makecell[l]{Bai Yun and her son were \\ returned to China in 2019.} & Rejected \\
    15 & 2024.12.02 & \makecell[l]{Who was fatally shot outside a New York City \\ hotel early on December 4, 2024?} & \makecell[l]{UnitedHealthcare CEO Brian Thompson \\ was fatally shot outside a New York City hotel \\ early on December 4, 2024.} & Accepted \\
    16 & 2025.09.10 & \makecell[l]{What did Erika Kirk say to her husband's killer \\ during her statement on September 12, 2025?} & \makecell[l]{Erika Kirk said to her husband's killer: \\ “you have no idea of the fire you have ignited.”} & Rejected \\
    17 & 2021.04.30 & \makecell[l]{On what date in 2021 did \\ Cinco de Mayo occur?} & May 18, 2021 & Rejected \\
    19 & 2021.05.28 & \makecell[l]{What was the name of the Duke University \\ basketball coach whose career was highlighted \\ in the article published on June 2, 2021?} & Mike Krzyzewski & Rejected \\
    20 & 2021.12.24 & \makecell[l]{On December 24, 2021, which university \\ removed the 'Goddess of Democracy' statue \\ in Hong Kong?} & \makecell[l]{The Chinese University of Hong Kong \\ removed the 'Goddess of Democracy' statue \\ on December 24, 2021.} & Rejected \\
    \hline
  \end{tabular}}
  \caption{Examples of benchmark question selection criteria. Accepted questions met unpredictability requirements; rejected questions were either answerable through general knowledge, referred to scheduled events, or contained other disqualifying characteristics. None of the questions shown were included in the final LLMLagBench dataset}
    \label{tab:questions_for_bench}
\end{table*}

Beyond identifying training cutoffs, LLMLagBench also provides an averaged overview of LLM performance on time-sensitive factual knowledge spanning 2021-2025. Table \ref{tab:model_summary} summarizes the results across all evaluated models, ranked by average faithfulness score. The table presents several key metrics: overall average faithfulness (0-2 scale), overall refusal rate, the number and dates of detected changepoints, mean faithfulness in the first segment (representing performance within the model's primary training period), and post-changepoint refusal rate (for the period after the last changepoint).

This ranking reveals substantial variation in model capabilities, with top-performing models like xAI's Grok-4 (average faithfulness 1.41) and Grok-3 (1.37) demonstrating significantly stronger knowledge of recent events compared to lower-ranked models. The table also illustrates the diversity of cutoff patterns: while most models exhibit one or two changepoints, some models like Mixtral-8x22B show three distinct boundaries, and Qwen2.5-Omni-7B shows no detectable changepoint due to consistently low performance throughout the period. The post-changepoint refusal rates vary significantly across models, from as low as 13\% (Gemma 3-4B) to 99\% (Gemini 2.5-Flash), reflecting different post-training behavior regarding knowledge uncertainty.

\section{Limitations}
The reported cutoff estimates are subject to two sources of error. First, despite rigorous manual curation, some questions may remain answerable through inference or general knowledge rather than requiring specific event knowledge. Consider the question: "On September 4, 2025, who won the US Open women's singles semi-final match between Aryna Sabalenka and Jessica Pegula?" With only two possible outcomes, a model has a 50\% chance of guessing correctly without any knowledge of the actual event, introducing measurement noise.

Second, the evaluator LLM may occasionally assign high faithfulness scores to plausible but incorrect responses. For instance, when asked "On what date did George Foreman die?" (correct answer: March 21, 2025), a model responding "George Foreman is still alive as of 2025" provides a reasonable statement based on its training data, which could receive a high faithfulness score despite being factually incorrect. Such evaluation errors are evidenced by occasional maximum scores (faithfulness = 2) appearing after clear cutoff points in our results.

Despite these error sources, their impact on our findings is limited. To illustrate, we examined pllum-24b-nc-chat-dpo-1507, a continually pretrained version of mistral-small-3.2-24b-instruct. Among 252 post-cutoff questions, this model achieved only 20 maximum scores (7.9\%), with the remaining 232 questions scoring 1 (10.7\%) or 0 (81.4\%). These isolated high scores resulting from guessing or evaluator errors did not prevent PELT from correctly identifying the knowledge boundary.

The comparison with the original mistral-small-3.2-24b-instruct is particularly informative: this base model consistently refused nearly all questions about events from 2025, attempting only 15 answers (all incorrect). This demonstrates that stronger instruction-tuning may yield lower absolute scores beyond the cutoff through more persistent refusal behavior. However, both patterns---consistent low scores and consistent refusals---represent the systematic shifts that PELT reliably detects.

\section{Conclusions and Future Work}

We introduced LLMLagBench, a benchmark designed to empirically identify temporal knowledge boundaries in LLMs through systematic evaluation of their performance on dated factual events. Our findings demonstrate a number of important insights.

First, knowledge infusion operates differently across training phases resulting in multiple partial cutoff points within a single model rather than a single sharp boundary. Second, smaller models show substantially weaker performance on time-sensitive questions and lower post-cutoff refusal rates, making them more susceptible to hallucination about recent events. Third, provider-declared cutoffs and model self-reports may diverge from empirically detected boundaries, with discrepancies ranging from months to years, underscoring the necessity of independent empirical validation.

In our future work we plan to extend LLMLagBench to evaluate retention of localized knowledge by developing country-specific versions using regional news sources, beginning with Polish news reports. Current version of LLMLagBench will be evaluated on a regular basis.

\section{Availability}

LLMLagBench is publicly available as an interactive leaderboard at \href{https://huggingface.co/spaces/pelcra/llmlagbench}{https://huggingface.co/spaces/pelcra/llmlagbench}. The underlying dataset of time-sensitive questions is withheld to prevent leaks.   Model developers and researchers can request evaluations by contacting \href{mailto:pelcra@uni.lodz.pl}{pelcra@uni.lodz.pl}.

\section*{Acknowledgments}

This work was supported by the CLARIN-BIZ-bis grant (FENG.02.04-IP.04-0004/24).

\bibliography{main_text}

\begin{thebibliography}{60}
\providecommand{\natexlab}[1]{#1}

\bibitem[{AGI et~al.(2025)AGI, Langford, Shah, Gupta, Bhatter, Goyal, Mathur,
  Mohanty, Kumar, Sethi, Komma, Pena, Jain, Kunysz, Opyrchal, Singh, Rawal,
  Prasad, Gispert, Kumar, Aryamane, Nair, M, Iyengar, Shanbhogue, He, Cervone,
  Loeb, Zhang, Fu, Lisnichenko, Zhipa, Potamianos, Kebarighotbi, Daronkolaei,
  Parmesh, Samra, Khan, Rez, Saffari, Agarwalla, Jhindal, Mamidala, Asmro,
  Ballakur, Mishra, Sridharan, Dubinina, Lenz, Doerr, Keating, Leaver, Smith,
  Wirth, Davey, Rosenbaum, Sohn, Chan, Chakrabarti, Ramakrishna, Roy, Iyer,
  Narayan-Chen, Yennu, Dabrowska, Gawlowska, Rumshisky, Turek, Deoras,
  Bezruchkin, Prasad, Dewan, Kiran, Gupta, Galstyan, Manoharan, Biswas, Mandal,
  Gupta, Pathan, Nagarajan, Rajasekaram, Sundararajan, Ganesan, Swaminathan,
  Mouchtaris, Champeau, Ray, Jaiswal, Sharma, Keefer, Muthiah, Leon-Millan,
  Koopman, Li, Biggs, Ott, Vinzamuri, Venkatesh, Ganesh, Vasani, Byrne, Hsu,
  Wang, King, Gorny, Feng, Zheng, Paul, Sun, Luo, Chen, Xie, Yu, Jugan, Panosh,
  Collins, Thompson, Karakus, Liu, Lambrecht, Lin, Wang, Yuan, Loyda, Walczak,
  Choppa, Prakash, Meas, Peris, Recaido, Xu, Sharma, Kernan, Thanapirom, Su,
  Xu, Yin, Ye, Tao, Parameshwara, Chang, Li, Hench, Tran, Dupuy, Davis,
  DiPersio, Christodoulopoulos, Li, Chen, Bovi, Chung, Hawkins, Harris, Ropell,
  He, Joo, Hwang, Rosen, Elkind, Pressel, Zhang, Kimball, Sorokin, Goodell,
  Modolo, Zhu, Suresh, Ragha, Filimonov, Kune, Rodriguez, Hazarika, Ram,
  Parkar, Patel, Desai, Rajput, Sule, Singh, Genzel, Goldenberg, He, Hanciu,
  Tharmal, Siankovich, Cikovic, Abraham, Sabir, Olson, Steven, Barut, Jackson,
  Wu, Chen, Mahalingam, Triefenbach, Yang, Liu, Wu, Tavakoli, Khozeimeh, Niu,
  Hieber, Li, Elbey, Krebs, Saupe, Sprünken, Fan, Khan, Vincenzo, Kang, Ding,
  He, Yeung, Qaddoumi, Karamanolakis, Huybrechts, Maddali, Iglesias, McShane,
  Sahin, Huang, Kwon, Sigurdsson, Chadha, Kosuru, Fuerstenau, Hah, Maideen,
  Hosokawa, Liu, Hsu, Wang, Li, Yang, Zhu, Fan, Singh, Kaluvala, Saeed, Xie,
  Feng, Luo, Pei, Nielsen, Ilati, Patel, Li, Lin, Raza, Cullinan, Kiss,
  Thangamani, Fadnavis, Sorodoc, Ertuerk, Yemialyanava, Soni, Jelal, Tse,
  FitzGerald, Zhao, Rothgeb, Lee, Jung, Debski, Tomczak, Jeun, Sanders,
  Crowley, Lee, Paidy, Tiwari, Farmer, Solinsky, Lau, Savareese, Zagorski, Dai,
  {Jiacheng}, {Gu}, Li, {Jian}, {Zheng}, Lu, Wang, Dai, Mo, Xu, Liang, Yang,
  Logan, Majmudar, Liu, Miao, Yi, Jin, Kao, Wang, Wang, Pemberton, Carlson,
  Blundell, Chin-Jew, He, Ho, Hueser, Lunt, Lee, Tan, Chatterjee, Gaspers,
  Wang, Fang, Tang, Wan, Wu, Wang, Shi, Chiu, Satriano, Yee, Dhamala, Bansal,
  Zhen, Chang, Lin, Raman, Sathyendra, Moroe, Bhandarkar, Kothari, Owczarzak,
  Gopalswamy, Ravi, Ramakrishnan, Arumugam, Mehta, Konczalska, Ravikumar, Tran,
  Qin, Li, Li, Kulkarni, Rodrigues, Patel, Abboud, Hajebi, Reiter, Schultz,
  Anisetty, Kotnana, Li, Channamallikarjuna, Jakubczyk, Pierewoj, Pal,
  Srivastav, Bannerman, Poddar, Prasad, Tseng, Naik, Vankadara, Minorics, Liu,
  Lausen, Ribeiro, Zhang, Gehorsam, Qi, Bauer, Knapp, Zeng, Tong, Wong, Chen,
  Rudnicki, Namazifar, Jaliminche, Tanke, Gupta, Ahlawat, Khanuja, Sundaram,
  Leyk, Momotko, Boese, Dreyer, Mueller, Fu, Górski, Mastalerczyk, Mora,
  Johnson, Scott, Wen, Barysau, Boumerdassi, Krishnan, Gupta, Hirani, Kulkarni,
  Narayanasamy, Bradford, Gens, Burke, Jin, Chen, Denkowski, Heymel,
  Krestyaninov, Obirek, Wichorowska, Miotk, Watroba, Hong, Yu, Liu, Gouda,
  El-Shabani, Ghavamzadeh, Bansal, Ziyadi, Xia, Susanj, Bhasin, Goswami,
  Belgamwar, Anastassacos, Bergeron, Jain, Jain, Chopparapu, Xu, Strom,
  Malandrakis, Mishra, Parkhi, Mehrabi, Sant, Gupta, Sekhar, Rajeev,
  Chidambaram, Dhar, Bhagwagar, Konforty, Babu, Razavi, Majumder, Dar, Hsu,
  Kvitca, Pandey, Seegmiller, Lange, Ferraro, Motwani, Kharazmi, Wang, Liu,
  Bradtke, Götz, Zhou, Wang, Poskart, Sonawane, Natarajan, Ramadorai, Shah,
  Nirantar, Chavali, Wanigasekara, Saraf, Dey, Pant, Pradhan, Patel, Dadlani,
  Sadha, Dong, Hu, {Qiaozi}, {Gao}, Liu, Lam, Do, Manmatha, Willis, Liu,
  Ellert, Kalinski, Attrach, Prasad, Prasad, Kunani, Gupta, Sharma, Tewari,
  Baskaran, Singh, Gupta, Reddy, Das, Chada, Mahesh, Chandrasekaran, Nallapati,
  Xue, Gangadharaiah, Rachakonda, Zhang, Blloshmi, Agrawal, Enyedi, Lowe,
  Shrestha, Piramuthu, Asad, Khanna, Mukherjee, Mittal, Prasad, Kumar, Diamant,
  Gupta, Li, Li, Fegade, Zhang, Arbow, Chen, Gabbard, Hoium, King, Iyer,
  Malick, Movaghati, Balakavi, Jakka, Paruvelli, Jayanthi, Mujumdar, Kapoor,
  Beygi, Dingliwal, Soltan, Ricklin, Tucker, Sinha, Choudhary, Tan, Broscheit,
  Schulter, Agarwal, Atluri, Valstar, Shankar, Sanyukta, Khanna, Khetrapal,
  Janakiraman, Shah, Akolkar, Giri, Khandelwal, Pawar, Sahu, Huang, Ra, Gopal,
  Dobroshinsky, Saba, Roy, Lal, Ananthakrishnan, Li, Srijan, Bhide, Tang, Zha,
  Oraby, Mostafa, Li, Bharathi, Prakash, Huang, Yembarwar, Pansare,
  Subramanian, Joshi, Liu, Tang, Chandak, Garg, Katiyar, Mehta, Srivastav,
  Yang, S, Choudhary, Senger, Babb, Moeini, Deng, Loganathan, Domagala, Narkar,
  Wadhwa, Zhang, Jiang, Trenous, Sarkar, Saha, Reddy, Dokania, Sandiri,
  Matsoukas, Bodapati, Wdaru, Venkateshdatta, Ronanki, Veeravanallur,
  Venkatapathy, Sankaraguru, Gorantla, Karuturi, Schroedl, Rongali, Kundu,
  Shakiah, Tiwari, Bharti, Sami, Mathew, Yu, Kim, Malode, Riel, Palod, Roy,
  Furqhan, Chung, Yoshitani, Yang, Chillakura, Bajwa, Lajumoke, Tran, Gueudre,
  Jung, Li, Seemman, Leffel, Xiang, Patel, Domhan, Falke, Guo, Li, Horszczaruk,
  Jedynak, Kulkarni, Marin, Metrycki, Wang, Jain, Singh, Chirimar, Gupta, Shah,
  Deshpande, Gunjal, Srikeshava, Vivek, Bharadwaj, Gangal, Kumar, Elango,
  Ordonez, Soto, Radhakrishnan, Patel, Singh, Kolanuvada, Kumar, Auvray,
  Cartillier, Ponzo, Peng, Khandelwal, Naik, Sahasrabudhe, Korolev, Gokuladas,
  Madan, Subramanian, Cevher, Gupta, Hamza, Zhang, Ruan, Cheng, Zhang, Zhao,
  Yao, Ouyang, Dashner, Campbell, Lin, Martin, Pearson, Jiang, Lu, Shi, Peng,
  Gao, Jiang, Fei, Wang, Zhou, Feng, Zhao, Wang, Li, Zhang, Wang, Fu, Yuan,
  Wang, Rao, Tavizon, Rossiytsev, Chen, Liu, Zou, Park, Versley, Zhang, Patel,
  Lu, Pan, {Yi-Hsiang}, {Lai}, Hu, Wang, Zhou, Xiang, Shi, Wang, Galatzer,
  Wang, Shen, Sun, Purwatama, {Yue}, {Wu}, Gu, Wang, Zeng, Chen, Zhou, Xie,
  Guy, Ambrozinski, Cai, Zhang, Wang, Jin, Zhao, Li, Luo, Zhang, Fang, Bu,
  Wang, Li, Wang, {Zimeng}, {Qiu}, and Li}]{agi_amazon_2025}
Amazon AGI, Aaron Langford, Aayush Shah, Abhanshu Gupta, Abhimanyu Bhatter,
  Abhinav Goyal, Abhinav Mathur, Abhinav Mohanty, Abhishek Kumar, Abhishek
  Sethi, Abi Komma, Abner Pena, Achin Jain, Adam Kunysz, Adam Opyrchal, Adarsh
  Singh, Aditya Rawal, Adok Achar~Budihal Prasad, Adrià~de Gispert, and 767
  others. 2025.
\newblock The {Amazon} {Nova} {Family} of {Models}: {Technical} {Report} and
  {Model} {Card}.
\newblock \emph{arXiv preprint}.
\newblock ArXiv:2506.12103.
\newblock URL: \url{http://arxiv.org/abs/2506.12103}, \href
  {https://doi.org/10.48550/arXiv.2506.12103}
  {\path{doi:10.48550/arXiv.2506.12103}}.

\bibitem[{AI21Labs()}]{jamba-hf}
AI21Labs.
\newblock ai21labs/{AI}21-{J}amba-{L}arge-1.7.
\newblock URL: \url{https://huggingface.co/ai21labs/AI21-Jamba-Large-1.7}.

\bibitem[{Aminikhanghahi and Cook(2017)}]{aminikhanghahi2017survey}
Samaneh Aminikhanghahi and Diane~J Cook. 2017.
\newblock A survey of methods for time series change point detection.
\newblock \emph{Knowledge and Information Systems}, 51(2):339--367.

\bibitem[{Anthropic({\natexlab{a}})}]{claude-3}
Anthropic. {\natexlab{a}}.
\newblock The {C}laude 3 {M}odel {F}amily: {O}pus, {S}onnet, {H}aiku.
\newblock URL:
  \url{https://www-cdn.anthropic.com/de8ba9b01c9ab7cbabf5c33b80b7bbc618857627/Model_Card_Claude_3.pdf}.

\bibitem[{Anthropic({\natexlab{b}})}]{claude-4}
Anthropic. {\natexlab{b}}.
\newblock Models overview.
\newblock URL:
  \url{https://docs.claude.com/en/docs/about-claude/models/overview}.

\bibitem[{Anthropic(2024{\natexlab{a}})}]{claude-3-haiku-release}
Anthropic. 2024{\natexlab{a}}.
\newblock Claude 3 {H}aiku: our fastest model yet.
\newblock URL: \url{https://www.anthropic.com/news/claude-3-haiku}.

\bibitem[{Anthropic(2024{\natexlab{b}})}]{claude-3-opus-release}
Anthropic. 2024{\natexlab{b}}.
\newblock Introducing the next generation of {C}laude.
\newblock URL: \url{https://www.anthropic.com/news/claude-3-family}.

\bibitem[{Anthropic(2024{\natexlab{c}})}]{claude-3.5}
Anthropic. 2024{\natexlab{c}}.
\newblock Model {C}ard {A}ddendum: {C}laude 3.5 {H}aiku and {U}pgraded {C}laude
  3.5 {S}onnet.
\newblock URL:
  \url{https://assets.anthropic.com/m/1cd9d098ac3e6467/original/Claude-3-Model-Card-October-Addendum.pdf}.

\bibitem[{Anthropic(2025)}]{claude-4-release}
Anthropic. 2025.
\newblock Introducing {C}laude 4.
\newblock URL: \url{https://www.anthropic.com/news/claude-4}.

\bibitem[{Cheng et~al.(2024)Cheng, Marone, Weller, Lawrie, Khashabi, and
  Durme}]{cheng2024dateddatatracingknowledge}
Jeffrey Cheng, Marc Marone, Orion Weller, Dawn Lawrie, Daniel Khashabi, and
  Benjamin~Van Durme. 2024.
\newblock Dated data: Tracing knowledge cutoffs in large language models.
\newblock \emph{Preprint}, arXiv:2403.12958.
\newblock URL: \url{https://arxiv.org/abs/2403.12958}, \href
  {https://arxiv.org/abs/2403.12958} {\path{arXiv:2403.12958}}.

\bibitem[{Chu et~al.(2024)Chu, Chen, Chen, Yu, Wang, Liu, and
  Qin}]{chu_timebench_2024}
Zheng Chu, Jingchang Chen, Qianglong Chen, Weijiang Yu, Haotian Wang, Ming Liu,
  and Bing Qin. 2024.
\newblock {TimeBench}: {A} {Comprehensive} {Evaluation} of {Temporal}
  {Reasoning} {Abilities} in {Large} {Language} {Models}.
\newblock In \emph{Proceedings of the 62nd {Annual} {Meeting} of the
  {Association} for {Computational} {Linguistics} ({Volume} 1: {Long}
  {Papers})}, pages 1204--1228, Bangkok, Thailand. Association for
  Computational Linguistics.
\newblock URL: \url{https://aclanthology.org/2024.acl-long.66/}, \href
  {https://doi.org/10.18653/v1/2024.acl-long.66}
  {\path{doi:10.18653/v1/2024.acl-long.66}}.

\bibitem[{Cohere({\natexlab{a}})}]{cohere_command_2025}
Cohere. {\natexlab{a}}.
\newblock Cohere's {C}ommand {A} {M}odel.
\newblock URL: \url{https://docs.cohere.com/docs/command-a}.

\bibitem[{Cohere({\natexlab{b}})}]{command-r}
Cohere. {\natexlab{b}}.
\newblock Cohere's {C}ommand {R} {M}odel.
\newblock URL: \url{https://docs.cohere.com/docs/command-r}.

\bibitem[{Cohere({\natexlab{c}})}]{cohere_command_r}
Cohere. {\natexlab{c}}.
\newblock Cohere's {C}ommand {R}+ {M}odel.
\newblock URL: \url{https://docs.cohere.com/docs/command-r-plus}.

\bibitem[{Cohere(2025)}]{command-a-release}
Cohere. 2025.
\newblock Announcing {C}ommand {A}.
\newblock URL: \url{https://docs.cohere.com/changelog/command-a}.

\bibitem[{Comanici et~al.(2025)Comanici, Bieber, Schaekermann, Pasupat,
  Sachdeva, Dhillon, Blistein, Ram, Zhang, Rosen, Marris, Petulla, Gaffney,
  Aharoni, Lintz, Pais, Jacobsson, Szpektor, Jiang, Haridasan, Omran, Saunshi,
  Bahri, Mishra, Chu, Boyd, Hekman, Parisi, Zhang, Kawintiranon, Bedrax-Weiss,
  Wang, Xu, Purkiss, Mendlovic, Deutel, Nguyen, Langley, Korn, Rossazza, Ramé,
  Waghmare, Miller, Byrd, Sheshan, Bhardwaj, Janus, Rissa, Horgan, Silver,
  Wahid, Brin, Raimond, Kloboves, Wang, Gundavarapu, Shumailov, Wang,
  Pajarskas, Heyward, Nikoltchev, Kula, Zhou, Garrett, Kafle, Arik, Goel, Yang,
  Park, Kojima, Mahmoudieh, Kavukcuoglu, Chen, Fritz, Bulyenov, Roy, Paparas,
  Shemtov, Chen, Strudel, Reitter, Roy, Vlasov, Ryu, Leichner, Yang, Mariet,
  Vnukov, Sohn, Stuart, Liang, Chen, Rawlani, Koh, Co-Reyes, Lai, Banzal,
  Vytiniotis, Mei, Cai, Badawi, Fry, Hartman, Zheng, Jia, Keeling, Louis, Chen,
  Robles, Hung, Zhou, Saxena, Goenka, Ma, Fisher, Taege, Graves, Steiner, Li,
  Nguyen, Sukthankar, Stanton, Eslami, Shen, Akin, Guseynov, Zhou, Alayrac,
  Joulin, Farkash, Thapliyal, Roller, Shazeer, Davchev, Koo, Forbes-Pollard,
  Audhkhasi, Farquhar, Gilady, Song, Aslanides, Mendolicchio, Parrish, Blitzer,
  Gupta, Ju, Yang, Datta, Tacchetti, Mehta, Dibb, Gupta, Piccinini, Hadsell,
  Rajayogam, Jiang, Griffin, Sundberg, Hayes, Frolov, Xie, Zhang, Dasgupta,
  Kalra, Shani, Macherey, Huang, MacDermed, Duddu, Zacchello, Yang, Lo, Hui,
  Kastelic, Gasaway, Tan, Yue, Barrio, Wieting, Yang, Nystrom, Demmessie,
  Levskaya, Viola, Tekur, Billock, Necula, Joshi, Schaeffer, Lokhande, Sorokin,
  Shenoy, Chen, Collier, Li, Bos, Wichers, Lee, Pouget, Thangaraj, Axiotis,
  Crone, Sterneck, Chinaev, Krakovna, Ferludin, Gemp, Winkler, Goldberg,
  Korotkov, Xiao, Mehrotra, Mariserla, Piratla, Thurk, Pham, Ma, Senges, Kumar,
  Meyer, Talius, Pierse, Sandhu, Toma, Lin, Nath, Stone, Sadigh, Gupta, Guez,
  Singh, Thomas, Duerig, Gong, Tanburn, Zhang, Dao, Hammad, Xie, Rijhwani,
  Murdoch, Kim, Thompson, Cheng, Sohn, Sprechmann, Xu, Tadepalli, Young, Zhang,
  Srinivasan, Aperghis, Ayyar, Fitoussi, Burnell, Madras, Dusenberry, Xiong,
  Oguntebi, Albrecht, Bornschein, Mitrović, Dimarco, Shamanna, Shah, Sezener,
  Upadhyay, Lacey, Schiff, Baur, Ganapathy, Schnider, Wirth, Schenck,
  Simanovsky, Tan, Fränken, Duan, Mankalale, Dhawan, Sequeira, Wei, Goel,
  Unlu, Zhu, Sun, Balashankar, Shuster, Umekar, Alnahlawi, Oord, Chen, Zhai,
  Dai, Lee, Doi, Zilka, Vallu, Shrivastava, Lee, Husain, Zhuang, Cohen-Addad,
  Barber, Atwood, Sadovsky, Wellens, Hand, Rajendran, Turker, Carey, Xu,
  Soltau, Li, Song, Li, Kemaev, Brown, Burns, Patraucean, Stanczyk,
  Aravamudhan, Blondel, Noga, Blanco, Song, Isard, Sharma, Hayes, Badawy, Lamp,
  Laish, Kozlova, Chan, Singla, Sunkara, Upadhyay, Liu, Bai, Wilkiewicz,
  Zlocha, Liu, Li, Li, Barak, Raboshchuk, Choi, Liu, Jue, Sharma, Marzoca,
  Busa-Fekete, Korsun, Elisseeff, Shen, Carthy, Lamerigts, Hosseini, Lin, Chen,
  Yang, Chauhan, Omernick, Jia, Zainullina, Hassabis, Vainstein, Amid, Zhou,
  Votel, Vértes, Li, Zhou, Lazaridou, McMahan, Narayanan, Soyer, Basu, Lee,
  Perozzi, Cao, Berrada, Arya, Chen, {Katrina}, {Xu}, Lochbrunner, Hofer,
  Sharifzadeh, Wu, Goldman, Awasthi, Wang, Wu, Sha, Zhang, Mikuła, Graziano,
  Mcloughlin, Giannoumis, Namiki, Malik, Radebaugh, Hall, Leal-Cavazos, Chen,
  Sindhwani, Kao, Greene, Griffith, Welty, Montgomery, Yoshino, Yuan, Goodman,
  Michaely, Lee, Sawhney, Chen, Zheng, Shum, Savinov, Pot, Pak, Zadimoghaddam,
  Bhatnagar, Lewenberg, Kutzman, Liu, Katzen, Selier, Djolonga, Lepikhin, Xu,
  Liang, Tan, Schillings, Ersoy, Blois, Bandemer, Singh, Lebedev, Joshi, Brown,
  Palmer, Pathak, Jalan, Zubach, Lall, Parker, Gunjan, Rogulenko, Sanghai,
  Leng, Egyed, Li, Ivanova, Andriopoulos, Xie, Rosenfeld, Wright, Sharma, Geng,
  Wang, Kwei, Pan, Zhang, Wang, Liu, Yeung, Cole, Rosenberg, Yang, Chen,
  Polovets, Nair, Saxena, Smith, Chang, Mahendru, Grant, Iyer, Cai, McGiffin,
  Shen, Walton, Girgis, Woodman, Ke, Kwong, Rouillard, Rao, Li, Xu, Prost, Zou,
  Ji, Magni, Liechty, Calian, Ramachandran, Krivokon, Huang, Chen, Hauth,
  Ilić, Xi, Lim, Ion, Moradi, Toksoz-Exley, Bullard, Allamanis, Yang, Wang,
  Hong, Gergely, Li, Mittal, Kovalev, Ungureanu, Labanowski, Wassenberg,
  Lacasse, Cideron, Dević, Marsden, Nguyen, Fink, Zhong, Kiyono, Ivanov, Ma,
  Bain, Yalasangi, She, Petrushkina, Lunayach, Bromberg, Hodkinson, Meshram,
  Vlasic, Kyker, Xu, Stanway, Yang, Zhao, Tung, Odoom, Fujii, Gilmer, Kim,
  Halim, Le, Bohnet, El-Sayed, Neyshabur, Reynolds, Reich, Xu, Moreira, Sharma,
  Liu, Hosseini, Raisinghani, Su, Lao, Formoso, Gelmi, Gueta, Dey, Gribovskaya,
  Ćevid, Mudgal, Bingham, Wang, Kumar, Cullum, Han, Bousmalis, Cedillo, Chu,
  Magay, Michel, Hlavnova, Calandriello, Ariafar, Yao, Sehwag, Vezer, Lago,
  Zhu, Rubenstein, Porter, Baddepudi, Riva, Istin, Yeh, Li, Howard, Jha, Chen,
  Liedekerke, Ahmed, Rodriguez, Bhatia, Wang, Elqursh, Klinghoffer, Chen,
  Kohli, I, Zhang, Nado, Chen, Chen, Zhang, Singh, Hillier, Lebron, Tao, Liu,
  Dulac-Arnold, Zhang, Narayan, Liu, Firat, Bhowmick, Liu, Zhang, Zhang,
  Rotival, Howard, Sinha, Grushetsky, Beyret, Gopalakrishnan, Zhao, He,
  Payrits, Nabulsi, Zhang, Chen, Lee, Fallen, Gollapudi, Zhou, Pavetić,
  Köppe, Huang, Pasumarthi, Fernando, Fischer, Ćurko, Gao, Svensson, Stone,
  Qureshi, Sinha, Kulshreshtha, Matysiak, Mao, Saroufim, Faust, Duan, Fidel,
  Katircioglu, Kaufman, Shah, Kong, Bapna, Weisz, Dunleavy, Dutta, Liu,
  Chaabouni, Parada, Wu, Belias, Bissacco, Fort, Xiao, Huot, Knutsen, Blau, Li,
  Prendki, Love, Chow, Charoenpanit, Shimokawa, Coriou, Gregor, Izo, Akula,
  Pinto, Hahn, Paulus, Guo, Sharma, Hsieh, Chukwuka, Hashimoto, Rauschmayr, Wu,
  Angermueller, Wang, Gerlach, Pliskin, Mirylenka, Ma, Baugher, Gale, Bijwadia,
  Rakićević, Wood, Park, Chang, Seal, Tar, Krasowiak, Song, Stephanov, Wang,
  Maggioni, Lin, Wu, Paul, Jiang, Agrawal, Piot, Feng, Kim, Doshi, Lai,
  {Chuqiao}, {Xu}, Vikram, Chelba, Krause, Zhuang, Rae, Denk, Collister,
  Weerts, Luo, Lu, Garnes, Gupta, Spitz, Hassidim, Liang, Shafran, Humphreys,
  Vassigh, Wallis, Shejwalkar, Perez-Nieves, Hornung, Tan, Westberg, Ly, Zhang,
  Farris, Park, Kosik, Cankara, Maksai, Xu, Cassirer, Caelles, Abdolmaleki,
  Chiang, Fabrikant, Shetty, He, Giménez, Hashemi, Panthaplackel, Kulizhskaya,
  Deshmukh, Pighin, Alazard, Jindal, Noury, S, Qin, Dotiwalla, Spencer,
  Babaeizadeh, Chen, Mehta, Lees, Leach, Koanantakool, Akolzin, Comanescu, Ahn,
  Svyatkovskiy, Mustafa, D'Ambrosio, Garlapati, Lamblin, Agarwal, Song, Sessa,
  Coquinot, Maggs, Masoom, Pitta, Wang, Morris-Suzuki, Porter, Jia, Dudek, R,
  Paduraru, Ansell, Bolukbasi, Lu, Ganeshan, Wang, Griffiths, Benenson, He,
  Swirhun, Papamakarios, Chawla, Sengupta, Wang, Milutinovic, Mordatch, Jia,
  Smith, Ng, Nigam, Young, Vušak, Hechtman, Goenka, Zipori, Ayoub, Popat,
  Acharya, Yu, Bloxwich, Song, Roit, Li, Boag, Nayakanti, Chandra, Ding, Mehta,
  Hope, Zhang, Shtacher, Badola, Nakashima, Sozanschi, Comşa, Žužul,
  Caveness, Odell, Watson, Cesare, Lippe, Lockhart, Verma, Chen, Sun, Zhuo,
  Shah, Gupta, Muzio, Niu, Zait, Singh, Gaba, Ye, Ramachandran, Saleh, Popa,
  Dubey, Liu, Javanmardi, Epstein, Hemsley, Green, Ranka, Cohen, Fu, Ghemawat,
  Borovik, Martens, Chen, Shyam, Pinto, Yang, Ţifrea, Du, Gong, Agarwal, Kim,
  Frank, Shah, Song, Deng, Mikhalap, Chatziprimou, Chung, Creswell, Zhang, Jun,
  Lebsack, Truong, Andačić, Yona, Fornoni, Rong, Toropov, Soudagar, Audibert,
  Zaiem, Abbas, Rusu, Potluri, Weng, Kementsietsidis, Tsitsulin, Peng, Ha,
  Jain, Latkar, Ivanov, McLean, GP, Venkataraman, Liu, Krishnan, D'sa, Yogev,
  Collins, Lee, Ho, Doersch, Yona, Gao, Ferreira, Ozturel, Muckenhirn, Zheng,
  Balasubramaniam, Bansal, Driessche, Eiger, Haykal, Misra, Goyal, Martins,
  Leung, Valfridsson, Flynn, Bishop, Pang, Halpern, Yu, Moore, {Yuvein}, {Zhu},
  Thiagarajan, Drori, Xiao, Dery, Jagerman, Lu, Ge, Aggarwal, Khare, Tran,
  Elyada, Alet, Rubin, Chou, Tian, Bai, Chan, Lew, Misiunas, Bilal, Ray,
  Raghuram, Castro-Ros, Carpenter, Zheng, Kilgore, Broder, Xue, Kallakuri, Dua,
  Yuen, Chien, Schultz, Agrawal, Tsarfaty, Hu, Kannan, Marcus, Kothari, Sun,
  Horn, Bošnjak, Naeem, Hirsch, Chiang, Fang, Han, Wang, Hora, He, Lučić,
  Changpinyo, Tripathi, Youssef, Kwak, Schlattner, Graves, Leblond, Zeng,
  Andreassen, Rasskin, Song, Cao, Oh, Hoffman, Skut, Zhang, Stritar, Cai,
  Khanna, Wang, Sharma, Reisswig, Jun, Prasad, Sholokhova, Singh, Rosenthal,
  Ruoss, Beaufays, Kirmani, Chen, Schalkwyk, Herzig, Kim, Jacob, Vincent,
  Reyes, Balazevic, Hussenot, Schneider, Barnes, Castro, Babbula, Green, Cabi,
  Duduta, Driess, Galt, Velan, Wang, Jiao, Mauger, Phan, Patel, Galić, Chang,
  Marcus, Harvey, Salazar, Dabir, Sheth, Mandhane, Sedghi, Willcock, Zandieh,
  Prabhakara, Amini, Miech, Stone, Nicosia, Niemczyk, Xiao, Kim, Kwasiborski,
  Verma, Oflazer, Hirnschall, Sung, Liu, Everett, Bakker, Weisz, Wang,
  Sampathkumar, Shaham, Xu, Altun, Wang, Saeki, Chen, Taropa, Vasanth, Austin,
  Huang, Petrovic, Dou, Golovin, Rozhdestvenskiy, Culp, Wu, Sano, Jain,
  Proskurnia, Cevey, Ruiz, Patil, Mirzazadeh, Ni, Snaider, Fan, Fréchette,
  Pierigiovanni, Iqbal, Lee, Fantacci, Xing, Wang, Irpan, Raposo, Luan, Chen,
  Ganapathy, Hui, Nie, Guyon, Ge, Vij, Zheng, Lee, Castaño, Baatarsukh,
  Ibagon, Chronopoulou, FitzGerald, Viswanadha, Huda, Moroshko, Stoyanov,
  Kolhar, Vaucher, Watts, Kuncoro, Michalewski, Kambala, Batsaikhan, Andreev,
  Jurenka, Le, Chen, Jishi, Chakera, Chen, Kini, Yadav, Siddhant, Labzovsky,
  Lakshminarayanan, Bostock, Botadra, Anand, Bishop, Conway-Rahman, Agarwal,
  Donchev, Singhal, Quitry, Ponomareva, Agrawal, Ni, Krishna, Samsikova, Karro,
  Du, Glehn, Lu, Choquette-Choo, Qin, Zhang, Li, Tyam, Mishra, Lowe, Ji, Wang,
  Faruqui, Slone, Dalibard, Narayanaswamy, Lambert, Manzagol, Karliner, Bolt,
  Lobov, Kusupati, Ye, Yang, Zen, George, Bhutani, Lacombe, Riachi, Bansal,
  Soh, Gao, Yu, Yu, Nottage, Rojas-Esponda, Noraky, Gupta, Kotikalapudi, Chang,
  Deur, Graur, Mossin, Farnese, Figueira, Moufarek, Huang, Zochbauer, Ingram,
  Chen, Wu, Puigdomènech, Rechis, Yu, Padmanabhan, Zhu, Ko, Banino, Daruki,
  Selvan, Bhaswar, Diaz, Su, Scellato, Brennan, Han, Chung, Agrawal,
  Khandelwal, Sim, Lustman, Ritter, Guu, Xia, Jain, Wang, Hill, Rossini,
  Kostelac, Misiunas, Sabne, Kim, Iscen, Wang, Leal, Sreevatsa, Evci, Warmuth,
  Joshi, Suo, Lottes, Honke, Jou, Karp, Hu, Sahni, Taïga, Kong, Ghosh, Wang,
  Pavagadhi, Axelsson, Grigorev, Siegler, Lin, Wang, Parisotto, Maddineni,
  Subudhi, Ben-David, Pochernina, Keller, Avrahami, Yuan, Mehta, Liu, Yang,
  Kan, Lee, Funkhouser, Cheng, Shi, Sharma, Kelley, Eyal, Malkov, Tallec,
  Bahat, Yan, {Xintian}, {Wu}, Lindner, Wu, Caciularu, Luo, Jenatton, Zaman,
  Bi, Kornakov, Mallya, Ikeda, Karo, Singh, Evans, Netrapalli, Nallatamby,
  Tian, Assael, Raunak, Carbune, Bica, Madmoni, Cattle, Grover, Somandepalli,
  Lall, Vázquez-Reina, Patana, Mu, Talluri, Tran, Aggarwal, Skerry-Ryan, Xu,
  Burrows, Pan, Yvinec, Lu, Zhang, Nguyen, Mu, Barcik, Ran, Beltrone,
  Choromanski, Kharrat, Albanie, Purser-haskell, Bieber, Zhang, Wang, Hudson,
  Zhang, Fu, Mauerer, Bateni, Maschinot, Wang, Zhu, Pillai, Weyand, Liu,
  Akerlund, Bertsch, Premachandran, Jin, Roulet, Boursac, Mittal, Ndebele,
  Karadzhov, Ghalebikesabi, Liang, Wu, Cong, Ghelani, Singh, Fatemi, {Warren},
  {Chen}, Kwong, Kolganov, Li, Song, Kuang, Miryoosefi, Webster, Wendt, Socala,
  Su, Mendonça, Gupta, Li, Tsai, {Qiong}, {Hu}, Kang, Chen, Girgin, Xian, Lee,
  Ramsden, Baker, Elish, Krayvanova, Joshi, Simsa, Yang, Ambroszczyk, Ghosh,
  Kar, Shangguan, Yamamori, Akulov, Brock, Tang, Vashishtha, Munoz, Steiner,
  Andra, Eppens, Feng, Kobayashi, Goldshtein, Mahdy, Wang, {Jilei}, {Wang},
  Killam, Kwiatkowski, Kopparapu, Zhan, Jia, Bendebury, Luo, Recasens, Knight,
  Chen, Patel, Li, Withbroe, Weesner, Bhatia, Ren, Eisenbud, Songhori, Sun,
  Choma, Kementsietsidis, Manning, Roark, Farhan, Feng, Tatineni, Cobon-Kerr,
  Li, Hendricks, Noble, Breaux, Kushman, Peng, Xue, Tobin, Rogers, Lipschultz,
  Alberti, Vlaskin, Dehghani, Sharma, Warkentin, Lee, Uria, Juan, Chandorkar,
  Sheftel, Liu, Davoodi, Pigem, Dhamdhere, Ross, Hoech, Mahdieh, Liu, Li,
  McCafferty, Liu, Mircea, Song, Savant, Saade, Cherry, Hellendoorn, Goyal,
  Pucciarelli, Torres, Yahav, Lee, Sjoesund, Kirov, Chang, Ghoshal, Li,
  Baechler, Pereira, Sainath, Boral, Grewe, Halumi, Phu, Shen, Ribeiro, Varma,
  Kaskasoli, Feinberg, Potti, Kahn, Wisniewski, Mohamed, Hrafnkelsson,
  Shahriari, Lespiau, Patel, Yeung, Paine, Mei, Ramirez, Shivanna, Zhong,
  Woodward, Tubone, Khan, Chen, Nielsen, Ionescu, Prabhu, Gao, Wang,
  Augenstein, Subramaniam, Chang, Iliopoulos, Luo, Khan, Kuo, Teplyashin,
  Perot, Kilpatrick, Globerson, Yu, Siddiqui, Sukhanov, Kandoor, Gupta,
  Andreetto, Ambar, Kim, Wesołowski, Perrin, Limonchik, Fan, Stephan,
  Stewart-Binks, Kappedal, He, Cogan, Datta, Zhou, Ye, Kieliger, Ramalho,
  Kastner, Mentzer, Ko, Suggala, Zhou, Butt, Strejček, Belenki, Venugopalan,
  Ling, Eltyshev, Deng, Kovacs, Raghavachari, Dai, Schuster, Schwarcz, Nguyen,
  Nguyen, Buttimore, Mallick, Gandhe, Benjamin, Jastrzebski, Yan, Basu, Apps,
  Edkins, Allingham, Odisho, Kocisky, Zhao, Xue, Reddy, Anastasiou, Atias,
  Redmond, Milan, Heess, Schmit, Dafoe, Andor, Gangwani, Dragan, Zhang, Kachra,
  Wu, Xue, Aydin, Liu, Zhou, Malihi, Wu, Gopal, Schumann, Stys, Wang, Olšák,
  Liu, Schallhart, Mao, Brady, Xu, Mery, Sitawarin, Velusamy, Cobley, Zhai,
  Walder, Katz, Jawahar, Kulkarni, Yang, Paszke, Wang, Damoc, Borsos, Smith,
  Li, Gupta, Kapishnikov, Prakash, Luisier, Agarwal, Grathwohl, Chen, Han,
  Mehta, Over, Azizi, Meng, Santo, Zheng, Shapiro, Petrovski, Hui, Ghafouri,
  Snoek, Qin, Jordan, Sikora, Malmaud, Kuang, Świetlik, Sang, Shi, Li,
  Rosenberg, Zhao, Crawford, Peter, Lei, Garcia, Le, Wang, Amelot, Orr, Kacham,
  Alon, Tyen, Arora, Lyon, Kurakin, Ly, Guidroz, Yan, Panigrahy, Xu, Kagohara,
  Cheng, Noland, Lee, Lee, Yip, Wang, Nehoran, Bykovsky, Shan, Bhagatwala, Yan,
  Tan, Garrido, Ethier, Hurley, Vesom, Chen, Qiao, Nayyar, Walker, Sandhu,
  Rosca, Swisher, Dektiarev, Dillon, Muraru, Tragut, Myaskovsky, Reid, Velic,
  Xiao, George, Brand, Li, Yu, Gu, Deng, Aubet, Yeganeh, Alcober, Smith, Cohn,
  McKinney, Tschannen, Sampath, Cheon, Luo, Liu, Orbay, Peng, Botea, Zhang,
  Yoon, Magalhaes, Stradomski, Mackinnon, Hemingray, Venkatesan, May, Kim,
  Druinsky, Ye, Xu, Huang, Abdallah, Dostmohamed, Fellinger, Munkhdalai,
  Maurya, Garst, Zhang, Krikun, Bucher, Veerubhotla, Liu, Li, Gupta, Adamek,
  Chen, Orlando, Zaks, Amersfoort, Camp, Wan, Choe, Wu, Olszewska, Yu, Vadali,
  Scholz, Freitas, Lin, Hua, Liu, Ding, Zhou, Severson, Tsihlas, Yang, Spalink,
  Yerram, Pankov, Blevins, Vargas, Jauhari, Miecnikowski, Zhang, Kumar,
  Farabet, Lan, Flennerhag, Bitton, Ma, Bražinskas, Collins, Ahuja, Kudugunta,
  Bortsova, Giang, Zhu, Chi, Lundberg, Stern, Puttagunta, Xiong, Wu, Pande,
  Jhindal, Murphy, Clark, Brockschmidt, Deines, McKee, Bahir, Shen, Truong,
  McDuff, Gesmundo, Rosseel, Liang, Caluwaerts, Hamrick, Kready, Cassin,
  Ingale, Lao, Pollom, Ding, He, Bellot, Iljazi, Boppana, Han, Thompson,
  Khalifa, Bulanova, Mitrevski, Pang, Cooney, Shi, Coaguila, Yakar, Ranzato,
  Momchev, Rawles, Charles, Maeng, Zhang, Bansal, Zhao, Albert, Yuan,
  Vijayanarasimhan, Hirsch, Ramasesh, Vodrahalli, Wang, Gupta, Strouse, Ni,
  Patel, Taubman, Huo, Gharibian, Monteiro, Lam, Vasudevan, Chaudhary,
  Albuquerque, Gupta, Riedel, Hegde, Ruderman, György, Wainwright, Chaugule,
  Ayan, Levinboim, Shleifer, Kalley, Mirrokni, Rao, Radhakrishnan, Hartford,
  Wu, Zhu, Bertolini, Xiong, Serrano, Tomlinson, Ott, Chang, Graham, Li, Liang,
  Long, Borgeaud, Ahmad, Grills, Mincu, Izzard, Liu, Xie, O'Bryan, Ponda, Tong,
  Liu, Malkin, Salama, Chen, Anil, Rao, Swavely, Bilenko, Anderson, Tan, Xie,
  Wu, Yu, Vinyals, Ryabtsev, Dangovski, Baumli, Keysers, Wright, Ashwood, Chan,
  Shtefan, Guo, Bapna, Soricut, Pecht, Ramos, Wang, Cai, Trinh, Barham, Friso,
  Stickgold, Ding, Shakeri, Ardila, Briakou, Culliton, Raveret, Cui, Saxton,
  Roy, Azizi, Yin, Loher, Bunner, Choi, Ahmed, Li, Li, Dai, Elabd, Ganapathy,
  Agrawal, Hua, Kunkle, Rajayogam, Ahuja, Conmy, Vasiloff, Beak, Yew,
  Mudigonda, Wydrowski, Blanton, Wang, Dauphin, Xu, Polacek, Chen, Hu, Sho,
  Kunesch, Manshadi, Rutherford, Li, Hsiao, Barr, Tudor, Kecman, Nagrani,
  Pchelin, Sundermeyer, S, Karmarkar, Gao, Chole, Bachem, Gao, BC, Dibb,
  Verzetti, Hernandez-Campos, Lunts, Johnson, Trapani, Koster, Brusilovsky,
  Xiong, Mohabey, Ke, Zou, Sabolić, Campos, Palowitch, Morris, Qiu, Ponnuramu,
  Li, Sharma, Sodhia, Tekelioglu, Chuklin, Yenugula, Gemzer, Strinopoulos,
  El-Husseini, Wang, Zhong, Leurent, Natsev, Wang, Mahaarachchi, Zhu, Peng,
  Alabed, Lee, Brohan, Szlam, Oh, Kovsharov, Lee, Wong, Barnes, Thornton,
  Gimeno, Levy, Sevenich, Johnson, Mallinson, Dadashi, Wang, Ren, Lahoti, Dhar,
  Feldman, Zheng, Ulrich, Panait, Blokzijl, Baetu, Matak, Harlalka, Shah,
  Marian, Dincklage, Du, Ley-Wild, Brownfield, Schumacher, Stuken, Noghabi,
  Gupta, Ren, Malmi, Weissenberger, Huergo, Bauza, Lampe, Douillard,
  Seyedhosseini, Frostig, Ghahramani, Nguyen, Krishnakumar, Ye, Gupta, Nazari,
  Geirhos, Shaw, Eleryan, Damen, Palomaki, Xiao, Wu, Yuan, Meadowlark, Bilotti,
  Lin, Sridhar, Schroecker, Chung, Luo, Strohman, Liu, Zheng, Emond, Wang,
  Lampinen, Fukuzawa, Campbell-Ajala, Roy, Lee-Thorp, Wang, Naim, {Tony},
  {Nguy{\textbackslash}{\textasciitilde}ên}, Bensky, Gupta, Rogozińska, Fu,
  Pillai, Veličković, Drath, Neubeck, Tulsyan, Klimovskiy, Metzler, Stevens,
  Yeh, Yuan, Yu, Zhang, Go, Tsang, Xu, Wan, Galatzer-Levy, Sobell, Toki,
  Salesky, Zhou, Antognini, Douglas, Wu, Lelkes, Kim, Cavallaro, Salazar, Liu,
  Besley, Refice, Jia, Li, Sokolik, Kannan, Simon, Chick, Aharon, Gandhi,
  Daswani, Amiri, Birodkar, Ittycheriah, Grabowski, Chang, Sutton, {Zhixin},
  {Lai}, Telang, Sargsyan, Jiang, Hoffmann, Brichtova, Hessel, Halcrow, Jerome,
  Brown, Tomala, Buchatskaya, Yu, Menon, Moreno, Liao, Zayats, Tang, Mah,
  Shenoy, Siegman, Hadian, Kwon, Tu, Khajehnouri, Foley, Haghani, Wu, Keshava,
  Gupta, Bruguier, Yao, Karmon, Zintgraf, Wang, Piqueras, Jung, Brennan,
  Machado, Giustina, Tessler, Lee, Zhang, Moore, Daugaard, Frömmgen, Beattie,
  Zhang, Kasenberg, Geri, Qin, Tomar, Ouyang, Yu, Zhou, Mathews, Davis, Li,
  Gupta, Yates, Deng, Kemp, Joung, Vassilvitskii, Guo, LV, Dopson, Lachgar,
  McConnaughey, Choudhury, Dena, Cohen, Ainslie, Levi, Gopavarapu, Zablotskaia,
  Vallet, Bahargam, Tang, Tomasev, Dyer, Balle, Lee, Bono, Mendez, Zubov, Yang,
  Rendulic, Zheng, Hogue, Pundak, Leith, Bhoopchand, Han, Žanić, Schaul,
  Delakis, Iyer, Wang, Singh, Abdelhamed, Thomas, Brahma, Dib, Kumar, Zhou,
  Bai, Mishra, Sun, Anklin, Sukkerd, Agubuzu, Briukhov, Gulati, Sieb, Pardo,
  Nasso, Chen, Zhu, Sosea, Goldin, Rush, Hombaiah, Noever, Zhou, Haves, Phuong,
  Ades, Chen, Yang, Pagadora, Bileschi, Cotruta, Saputro, Pramanik, Ammirati,
  Garrette, Villela, Blyth, Akbulut, Jha, Rrustemi, Wongpanich, Nagpal, Wu,
  Rivière, Kishchenko, Srinivasan, Chen, Sinha, Pham, Jia, Hennigan, Bakalov,
  Attaluri, Garmon, Rodriguez, Wegner, Jia, Senter, Fiedel, Petek, Liu, Hardin,
  Lehri, Carreira, Smoot, Prasetya, Akazawa, Stefanoiu, Ho, Angelova, Lin, Kim,
  Chen, Sieniek, Li, Guo, Baltateanu, Tafti, Wunder, Olmert, Shukla, Shen,
  Kovelamudi, Venkatraman, Neel, Thoppilan, Connor, Benzing, Stjerngren,
  Ghiasi, Polozov, Howland, Weber, Chiu, Girirajan, Terzis, Wang, Li, Shalom,
  Tewari, Denton, Aharoni, Kalb, Zhao, Zhang, Filos, Rahtz, Jain, Fan,
  Rodrigues, Wang, Shin, Austin, Ring, Sanchez-Vargas, Hassen, Kessler, Alon,
  Zhang, Chen, Ma, Si, Hou, Mirhoseini, Wilson, Bacon, Roelofs, Shu, Vasudevan,
  Adler, Dwornik, Terzi, Lawlor, Askham, Bernico, Dong, Hidey, Kilgour, Liu,
  Bhupatiraju, Leonhard, Zuo, Talukdar, Wei, Severyn, Listík, Lee, Tripathi,
  Park, Matias, Liu, Ruiz, Jayaram, Tolins, Marcenac, Wang, Seybold, Prior,
  Sharma, Weber, Sirotenko, Sung, Du, Pavlick, Zinke, Freitag, Dylla, Arenas,
  Potikha, Goldman, Tao, Chhaparia, Voitovich, Dogra, Ražnatović, Tsai, You,
  Johnson, Tucker, Gu, Yoo, Majzoubi, Gabeur, Raad, Rhodes, Kolipaka, Howard,
  Sampemane, Li, Asawaroengchai, Nguyen, Zhang, Cour, Yu, Fu, Jiang, Huang,
  Surita, Iturrate, Karov, Collins, Baeuml, Fuchs, Shetty, Ramaswamy, Ebrahimi,
  Guo, Shar, Barth-Maron, Addepalli, Richter, Cheng, Rives, Zheng, Griesser,
  Dikkala, Zeldes, Safarli, Das, Srivastava, Khan, Li, Pandey, Markeeva, Belov,
  Yan, Rybiński, Chen, Nawhal, Quinn, Govindaraj, York, Roberts, Garg,
  Godbole, Abernethy, Das, Thiet, Tompson, Nham, Vats, Caine, Helmholz,
  Pongetti, Ko, An, Hu, Ling, Pawar, Leland, Kinoshita, Khawaja, Selvi, Ie,
  Sinopalnikov, Proleev, Tripuraneni, Bevilacqua, Lee, Sanford, Suh, Tran,
  Dean, Baumgartner, Heitkaemper, Gubbi, Toutanova, Xu, Thekkath, Rong, Jain,
  Xie, Virin, Li, Litchev, Powell, Bharti, Kraft, Hua, Ikonomidis, Hitron,
  Kumar, Matthey, Bridgers, Lax, Malhi, Skopek, Gupta, Cao, Rasquinha, Põder,
  Stokowiec, Roth, Li, Sander, Kessinger, Jain, Loper, Park, Yarom, Cheng,
  Guruganesh, Rao, Li, Barros, Sushkov, Ferng, Shah, Aharoni, Kumar, McConnell,
  Li, Wang, Pereira, Swanson, Jamil, Xiong, Vijayakumar, Shroff, Soparkar, Gu,
  Soares, Wang, Majmundar, Wei, Bailey, Kassner, Kawamoto, Žužić, Gomes,
  Gupta, Guzman, Dasgupta, Bai, Pan, Piccinno, Vogel, Ponce, Hutter, Chang,
  Jiang, Gog, Ionescu, Manyika, Pedregosa, Ragan, Behrman, Mullins, Devin,
  Pyne, Gawde, Chadwick, Gu, Tavakkol, Twigg, Goyal, Elue, Goldie,
  Venkatachary, Fei, Feng, Ritter, Leal, Dasari, Sun, Rochman, O'Donoghue, Liu,
  Sproch, Chen, Clay, Petrov, Sidhwani, Mihailescu, Panagopoulos, Piergiovanni,
  Bai, Powell, Karkhanis, Yacovone, Mitrichev, Kovac, Uthus, Yazdanbakhsh,
  Amos, Zheng, Zhang, Miao, Ramabhadran, Radpour, Thakoor, Newlan, Lang,
  Jankowski, Bharadwaj, Sarr, Ashraf, Mondal, Yan, Rawat, Velury, Kochanski,
  Eccles, Och, Sharma, Mahintorabi, Gurney, Muir, Cohen, Thakur, Bloniarz,
  Mujika, Pritzel, Caron, Rahman, Lang, Onoe, Sirkovic, Hoover, Jian, Duque,
  Narayanan, Soergel, Haig, Maggiore, Buch, Dean, Figotin, Karpov, Gupta, Zhou,
  Huang, Vaswani, Semturs, Shivakumar, Watanabe, Rajendran, Lu, Hou, Ye,
  Vashishth, Nti, Sakenas, Ni, DeCarlo, Bendersky, Bagri, Cano, Peake,
  Tokumine, Godbole, Guía, Lando, Selo, Ellis, Tarlow, Gillick, Epasto,
  Jonnalagadda, Wei, Xie, Taly, Paganini, Sundararajan, Toyama, Yu, Petrova,
  Pappu, Agrawal, Buthpitiya, Frye, Buschmann, Crocker, Tagliasacchi, Wang,
  Huang, Perel, Wieder, Kazawa, Wang, Cole, Gupta, Golan, Bang, Kulkarni,
  Franko, Liu, Reid, Dalmia, Whang, Cen, Sundaram, Ferret, Isik, Ionita, Sun,
  Shekhawat, Mohammad, Pham, Huang, Raman, Zhou, Mcilroy, Myers, Peng, Scott,
  Covington, Erell, Joshi, Oliveira, Noy, Nasir, Walker, Axelrod, Dozat, Han,
  Chu, Weinstein, Shukla, Chandrakaladharan, Poklukar, Li, Jin, Eruvbetine,
  Hansen, Dabush, Jacovi, Phatale, Zhu, Baker, Shomrat, Xiao, Pouget-Abadie,
  Zhang, Wei, Song, King, Huang, Zhu, Sun, Franco, Lin, Arora, {Hui}, {Li},
  Xia, Vilnis, Schain, Alarakyia, Prince, Phillips, Habtegebriel, Xu, Gui,
  Ontanon, Aroyo, Gill, Lu, Katariya, Madeka, Krishnan, Raghvendra, Freedman,
  Tay, Menghani, Choy, Shetty, Abolafia, Kukliansky, Chou, Lichtarge, Burke,
  Coleman, Guo, Jin, Bhattacharya, Langston, Li, Kotecha, Yakubovich, Chen,
  Petrov, Powell, He, Quick, Garg, Hwang, Lu, Bhojanapalli, Kjems, Mehran,
  Archer, Hasselt, Balakrishna, Kearns, Guo, Riesa, Sazanovich, Gao, Sauer,
  Yang, Sheng, Jimma, Gansbeke, Nikolaev, Wei, Millican, Zhao, Snyder, Bolelli,
  O'Brien, Xu, Xia, Yuan, Neelakantan, Barker, Yadav, Kirkwood, Ahmad, Wee,
  Grimstad, Wang, Wiethoff, Settle, Wang, Blundell, Chen, Duvarney, Hu,
  Ronneberger, Lee, Li, Chakladar, Butryna, Evangelopoulos, Desjardins,
  Kanerva, Wang, Nowak, Li, Loo, Khurshudov, Shafey, Baddi, Lenc, Razeghi,
  Lieber, Sinha, Ma, Su, Huang, Ushio, Klimczak-Plucińska, Mohamed, Chen,
  Osindero, Ginzburg, Lamprou, Bashlovkina, Tran, Khodaei, Anand, Di, Eskander,
  Vuyyuru, Liu, Kamath, Goldenberg, Bellaiche, Pluto, Rosgen, Mansoor, Wong,
  Ganesh, Bailey, Baird, Deutsch, Baek, Jia, Lee, Friesen, Braun, Lee, Panda,
  Hernandez, Williams, Liu, Liang, Autef, Pitler, Jain, Kirk, Bunyan, Elias,
  Yin, Reid, Pope, Putikhin, Samanta, Guadarrama, Kim, Rowe, Valentine, Yan,
  Salcianu, Silver, Song, Singh, Ye, DeBalsi, Merey, Ofek, Webson, Mourad,
  Kakarla, Lattanzi, Roy, Sluzhaev, Butterfield, Tonioni, Waters, Kopalle,
  Chase, Cohan, Rao, Berry, Voznesensky, Hu, Chiafullo, Chikkerur, Scrivener,
  Zheng, Wiesner, Macherey, Lillicrap, Liu, Walker, Welling, Davies, Huang,
  Ren, Shabat, Agostini, Iinuma, Zelle, Sathyanarayana, D'olimpio, Redshaw,
  Ginsberg, Murthy, Geller, Matejovicova, Chakrabarti, Julian, Chan, Hu,
  Jarrett, Agarwal, Challagundla, Li, Tata, Ding, Meng, Dai, Vezzani, Garg,
  Bulian, Jasarevic, Cai, Rajamani, Santoro, Hartmann, Liang, Perz, Jindal, Bu,
  Seo, Poplin, Goedeckemeyer, Ghazi, Khadke, Liu, Mather, Zhang, Shah, Chen,
  Wei, Shivam, Cao, Cho, Scarpati, Moffitt, Barbu, Jurin, Chang, Liu, Zheng,
  Dave, Kaeser-Chen, Yu, Abdagic, Gonzalez, Huang, Zhong, Schmid, Petrini,
  Wertheim, Zhu, Nguyen, Ji, Zhou, Zhou, Feng, Cohen, Rim, Phal, Georgiev,
  Brand, Ma, Li, Gupta, Wang, Dubov, Tarbouriech, Majumder, Li, Rink, Suman,
  Guo, Sun, Nair, Xu, Elhawaty, Cabrera, Han, Eisenschlos, Bai, Li, Bansal,
  Sellam, Khan, Nguyen, Mao-Jones, Parotsidis, Marcus, Fan, Zimmermann,
  Kochinski, Graesser, Behbahani, Caceres, Riley, Kane, Lefdal, Willoughby,
  Vicol, Wang, Zhang, Gill, Liang, Prasad, Mariooryad, Kazemi, Wang,
  Muralidharan, Voigtlaender, Zhao, Zhou, D'Souza, Mavalankar, Arnold, Young,
  Sarvana, Lee, Nasr, Zou, Kim, Haas, Patel, Bulut, Parkinson, Biles,
  Kalashnikov, To, Kumar, Austin, Greve, Zhang, Goel, Li, Yaroshenko, Chang,
  Jindal, Clark, Taitelbaum, Johnson, Roval, Ko, Mohananey, Schuler, Dodhia,
  Li, Osawa, Cui, Xu, Shah, Huang, Gruzewska, Clement, Verma, Sercinoglu, Qian,
  Shah, Yamaguchi, Modi, Kosakai, Strohmann, Zeng, Gunel, Qian, Tarango,
  Jastrzębski, David, Shan, Schuh, Lad, Gierke, Madhavan, Chen, Kurzeja,
  Santamaria-Fernandez, Chen, Cordell, Chervonyi, Garcia, Kannen, Perot, Ding,
  Cohen-Ganor, Lavrenko, Wu, Evans, Santos, Sewak, Brown, Hard, Puigcerver,
  Zheng, Liang, Gladchenko, Ingle, First, Sermanet, Magister, Velimirović,
  Reddi, Ricco, Agustsson, Adam, Levine, Gaddy, Holtmann-Rice, Wang, Sathe,
  Roy, Bratanič, Carin, Mehta, Bonacina, Cao, Finkelstein, Rieser, Wu,
  Altché, Scandinaro, Li, Vieillard, Sethi, Tanzer, Xing, Wang, Bhatia,
  Citovsky, Anthony, Lin, Shi, Jakobovits, Gibson, Apte, Lee, Chen, Byravan,
  Maniatis, Webster, Dai, Chen, Pan, Fadeeva, Gleicher, Luong, and
  Bhumihar}]{comanici_gemini_2025}
Gheorghe Comanici, Eric Bieber, Mike Schaekermann, Ice Pasupat, Noveen
  Sachdeva, Inderjit Dhillon, Marcel Blistein, Ori Ram, Dan Zhang, Evan Rosen,
  Luke Marris, Sam Petulla, Colin Gaffney, Asaf Aharoni, Nathan Lintz,
  Tiago~Cardal Pais, Henrik Jacobsson, Idan Szpektor, Nan-Jiang Jiang, and 3290
  others. 2025.
\newblock Gemini 2.5: {Pushing} the {Frontier} with {Advanced} {Reasoning},
  {Multimodality}, {Long} {Context}, and {Next} {Generation} {Agentic}
  {Capabilities}.
\newblock \emph{arXiv preprint}.
\newblock ArXiv:2507.06261.
\newblock URL: \url{http://arxiv.org/abs/2507.06261}, \href
  {https://doi.org/10.48550/arXiv.2507.06261}
  {\path{doi:10.48550/arXiv.2507.06261}}.

\bibitem[{Deepseek-AI(2025)}]{deepseek-3.1}
Deepseek-AI. 2025.
\newblock Deepseek-{V}3.1 {R}elease.
\newblock URL: \url{https://api-docs.deepseek.com/news/news250821}.

\bibitem[{{DeepSeek-AI} et~al.(2025{\natexlab{a}}){DeepSeek-AI}, Guo, Yang,
  Zhang, Song, Zhang, Xu, Zhu, Ma, Wang, Bi, Zhang, Yu, Wu, Wu, Gou, Shao, Li,
  Gao, Liu, Xue, Wang, Wu, Feng, Lu, Zhao, Deng, Zhang, Ruan, Dai, Chen, Ji,
  Li, Lin, Dai, Luo, Hao, Chen, Li, Zhang, Bao, Xu, Wang, Ding, Xin, Gao, Qu,
  Li, Guo, Li, Wang, Chen, Yuan, Qiu, Li, Cai, Ni, Liang, Chen, Dong, Hu, Gao,
  Guan, Huang, Yu, Wang, Zhang, Zhao, Wang, Zhang, Xu, Xia, Zhang, Zhang, Tang,
  Li, Wang, Li, Tian, Huang, Zhang, Wang, Chen, Du, Ge, Zhang, Pan, Wang, Chen,
  Jin, Chen, Lu, Zhou, Chen, Ye, Wang, Yu, Zhou, Pan, Li, Zhou, Wu, Ye, Yun,
  Pei, Sun, Wang, Zeng, Zhao, Liu, Liang, Gao, Yu, Zhang, Xiao, An, Liu, Wang,
  Chen, Nie, Cheng, Liu, Xie, Liu, Yang, Li, Su, Lin, Li, Jin, Shen, Chen, Sun,
  Wang, Song, Zhou, Wang, Shan, Li, Wang, Wei, Zhang, Xu, Li, Zhao, Sun, Wang,
  Yu, Zhang, Shi, Xiong, He, Piao, Wang, Tan, Ma, Liu, Guo, Ou, Wang, Gong,
  Zou, He, Xiong, Luo, You, Liu, Zhou, Zhu, Xu, Huang, Li, Zheng, Zhu, Ma,
  Tang, Zha, Yan, Ren, Ren, Sha, Fu, Xu, Xie, Zhang, Hao, Ma, Yan, Wu, Gu, Zhu,
  Liu, Li, Xie, Song, Pan, Huang, Xu, Zhang, and
  Zhang}]{deepseek-ai_deepseek-r1_2025}
{DeepSeek-AI}, Daya Guo, Dejian Yang, Haowei Zhang, Junxiao Song, Ruoyu Zhang,
  Runxin Xu, Qihao Zhu, Shirong Ma, Peiyi Wang, Xiao Bi, Xiaokang Zhang,
  Xingkai Yu, Yu~Wu, Z.~F. Wu, Zhibin Gou, Zhihong Shao, Zhuoshu Li, Ziyi Gao,
  and 181 others. 2025{\natexlab{a}}.
\newblock {DeepSeek}-{R1}: {Incentivizing} {Reasoning} {Capability} in {LLMs}
  via {Reinforcement} {Learning}.
\newblock \emph{arXiv preprint}.
\newblock ArXiv:2501.12948.
\newblock URL: \url{http://arxiv.org/abs/2501.12948}, \href
  {https://doi.org/10.48550/arXiv.2501.12948}
  {\path{doi:10.48550/arXiv.2501.12948}}.

\bibitem[{{DeepSeek-AI} et~al.(2025{\natexlab{b}}){DeepSeek-AI}, Liu, Feng,
  Xue, Wang, Wu, Lu, Zhao, Deng, Zhang, Ruan, Dai, Guo, Yang, Chen, Ji, Li,
  Lin, Dai, Luo, Hao, Chen, Li, Zhang, Bao, Xu, Wang, Zhang, Ding, Xin, Gao,
  Li, Qu, Cai, Liang, Guo, Ni, Li, Wang, Chen, Chen, Yuan, Qiu, Li, Song, Dong,
  Hu, Gao, Guan, Huang, Yu, Wang, Zhang, Xu, Xia, Zhao, Wang, Zhang, Li, Wang,
  Zhang, Zhang, Tang, Li, Tian, Huang, Wang, Zhang, Wang, Zhu, Chen, Du, Chen,
  Jin, Ge, Zhang, Pan, Wang, Xu, Zhang, Chen, Li, Lu, Zhou, Chen, Wu, Ye, Ye,
  Ma, Wang, Zhou, Yu, Zhou, Pan, Wang, Yun, Pei, Sun, Xiao, Zeng, Zhao, An,
  Liu, Liang, Gao, Yu, Zhang, Li, Jin, Wang, Bi, Liu, Wang, Shen, Chen, Zhang,
  Chen, Nie, Sun, Wang, Cheng, Liu, Xie, Liu, Yu, Song, Shan, Zhou, Yang, Li,
  Su, Lin, Li, Wang, Wei, Zhu, Zhang, Xu, Xu, Huang, Li, Zhao, Sun, Li, Wang,
  Yu, Zheng, Zhang, Shi, Xiong, He, Tang, Piao, Wang, Tan, Ma, Liu, Guo, Wu,
  Ou, Zhu, Wang, Gong, Zou, He, Zha, Xiong, Ma, Yan, Luo, You, Liu, Zhou, Wu,
  Ren, Ren, Sha, Fu, Xu, Huang, Zhang, Xie, Zhang, Hao, Gou, Ma, Yan, Shao, Xu,
  Wu, Zhang, Li, Gu, Zhu, Liu, Li, Xie, Song, Gao, and
  Pan}]{deepseek-ai_deepseek-v3_2025}
{DeepSeek-AI}, Aixin Liu, Bei Feng, Bing Xue, Bingxuan Wang, Bochao Wu, Chengda
  Lu, Chenggang Zhao, Chengqi Deng, Chenyu Zhang, Chong Ruan, Damai Dai, Daya
  Guo, Dejian Yang, Deli Chen, Dongjie Ji, Erhang Li, Fangyun Lin, Fucong Dai,
  and 181 others. 2025{\natexlab{b}}.
\newblock {DeepSeek}-{V3} {Technical} {Report}.
\newblock \emph{arXiv preprint}.
\newblock ArXiv:2412.19437.
\newblock URL: \url{http://arxiv.org/abs/2412.19437}, \href
  {https://doi.org/10.48550/arXiv.2412.19437}
  {\path{doi:10.48550/arXiv.2412.19437}}.

\bibitem[{GLM-4.5 et~al.(2025)GLM-4.5, Zeng, Lv, Zheng, Hou, Chen, Xie, Wang,
  Yin, Zeng, Zhang, Wang, Zhong, Liu, Lu, Cao, Zhang, Huang, Wei, Cheng, An,
  Niu, Wen, Bai, Du, Wang, Zhu, Zhang, Wen, Wu, Xu, Huang, Zhao, Cai, Yu, Li,
  Ge, Huang, Zhang, Xu, Zhu, Li, Yin, Lin, Yang, Jiang, Ai, Zhu, Wang, Pan,
  Wang, Sun, Li, Li, Hu, Zhang, Peng, Tai, Zhang, Wang, Yang, Liu, Zhao, Liu,
  Yan, Liu, Chen, Li, Zhao, Ren, Jiao, Zhao, Yan, Wang, Gui, Zhao, Liu, Li, Li,
  Lu, Wang, Yuan, Li, Du, Du, Liu, Zhi, Gao, Wang, Yang, Xu, Fan, Wu, Ding,
  Wang, Zhang, Li, Xu, Zhao, Zhai, Du, Dong, Lei, Tu, Yang, Lu, Li, Li,
  {Shuang-Li}, Yang, Yi, Yu, Tian, Wang, Yu, Tam, Liang, Liu, Wang, Jia, Gu,
  Ling, Wang, Fan, Pan, Zhang, Zhang, Fu, Zhang, Xu, Wu, Lu, Wang, Zhou, Pan,
  Zhang, Wang, Li, Su, Geng, Zhu, Yang, Li, Wu, Li, Liu, Wang, Li, Zhang, Liu,
  Yang, Zhou, Qiao, Feng, Liu, Zhang, Wang, Yao, Wang, Liu, Chai, Li, Zhao,
  Chen, Zhai, Xu, Huang, Wang, Li, Dong, and Tang}]{team_glm-45_2025}
GLM-4.5, Aohan Zeng, Xin Lv, Qinkai Zheng, Zhenyu Hou, Bin Chen, Chengxing Xie,
  Cunxiang Wang, Da~Yin, Hao Zeng, Jiajie Zhang, Kedong Wang, Lucen Zhong,
  Mingdao Liu, Rui Lu, Shulin Cao, Xiaohan Zhang, Xuancheng Huang, Yao Wei, and
  152 others. 2025.
\newblock {GLM}-4.5: {Agentic}, {Reasoning}, and {Coding} ({ARC}) {Foundation}
  {Models}.
\newblock \emph{arXiv preprint}.
\newblock ArXiv:2508.06471.
\newblock URL: \url{http://arxiv.org/abs/2508.06471}, \href
  {https://doi.org/10.48550/arXiv.2508.06471}
  {\path{doi:10.48550/arXiv.2508.06471}}.

\bibitem[{Google({\natexlab{a}})}]{gemma-3}
Google. {\natexlab{a}}.
\newblock Gemma 3 model card.
\newblock URL: \url{https://ai.google.dev/gemma/docs/core/model_card_3}.

\bibitem[{Google({\natexlab{b}})}]{gemma-release}
Google. {\natexlab{b}}.
\newblock Gemma releases.
\newblock URL: \url{https://ai.google.dev/gemma/docs/releases}.

\bibitem[{Google(2024)}]{gemini-2-release}
Google. 2024.
\newblock Introducing {G}emini 2.0: our new {AI} model for the agentic era.
\newblock URL:
  \url{https://blog.google/technology/google-deepmind/google-gemini-ai-update-december-2024/}.

\bibitem[{Holtermann et~al.(2025)Holtermann, Röttger, and
  Lauscher}]{holtermann2025world24hoursprobing}
Carolin Holtermann, Paul Röttger, and Anne Lauscher. 2025.
\newblock Around the world in 24 hours: Probing llm knowledge of time and
  place.
\newblock \emph{Preprint}, arXiv:2506.03984.
\newblock URL: \url{https://arxiv.org/abs/2506.03984}, \href
  {https://arxiv.org/abs/2506.03984} {\path{arXiv:2506.03984}}.

\bibitem[{Jiang et~al.(2024)Jiang, Sablayrolles, Roux, Mensch, Savary, Bamford,
  Chaplot, Casas, Hanna, Bressand, Lengyel, Bour, Lample, Lavaud, Saulnier,
  Lachaux, Stock, Subramanian, Yang, Antoniak, Scao, Gervet, Lavril, Wang,
  Lacroix, and Sayed}]{jiang_mixtral_2024}
Albert~Q. Jiang, Alexandre Sablayrolles, Antoine Roux, Arthur Mensch, Blanche
  Savary, Chris Bamford, Devendra~Singh Chaplot, Diego de~las Casas, Emma~Bou
  Hanna, Florian Bressand, Gianna Lengyel, Guillaume Bour, Guillaume Lample,
  Lélio~Renard Lavaud, Lucile Saulnier, Marie-Anne Lachaux, Pierre Stock,
  Sandeep Subramanian, Sophia Yang, and 7 others. 2024.
\newblock Mixtral of {Experts}.
\newblock \emph{arXiv preprint}.
\newblock ArXiv:2401.04088.
\newblock URL: \url{http://arxiv.org/abs/2401.04088}, \href
  {https://doi.org/10.48550/arXiv.2401.04088}
  {\path{doi:10.48550/arXiv.2401.04088}}.

\bibitem[{Killick et~al.(2012)Killick, Fearnhead, and Eckley}]{Killick_2012}
Rebecca Killick, Paul Fearnhead, and Idris~A. Eckley. 2012.
\newblock Optimal detection of changepoints with a linear computational cost.
\newblock \emph{Journal of the American Statistical Association},
  107(500):1590–1598.
\newblock URL: \url{http://dx.doi.org/10.1080/01621459.2012.737745}, \href
  {https://doi.org/10.1080/01621459.2012.737745}
  {\path{doi:10.1080/01621459.2012.737745}}.

\bibitem[{Kimi(2025)}]{kimi-release}
Kimi. 2025.
\newblock Kimi {K}2: {O}pen {A}gentic {I}ntelligence.
\newblock URL: \url{https://arxiv.org/pdf/2507.20534}.

\bibitem[{Li and Goyal(2025)}]{li-goyal-2025-memorization}
Aochong~Oliver Li and Tanya Goyal. 2025.
\newblock Memorization vs. reasoning: Updating {LLM}s with new knowledge.
\newblock In \emph{Findings of the Association for Computational Linguistics:
  ACL 2025}, pages 25853--25874, Vienna, Austria. Association for Computational
  Linguistics.
\newblock URL: \url{https://aclanthology.org/2025.findings-acl.1326/}, \href
  {https://doi.org/10.18653/v1/2025.findings-acl.1326}
  {\path{doi:10.18653/v1/2025.findings-acl.1326}}.

\bibitem[{Lieber et~al.(2024)Lieber, Lenz, Bata, Cohen, Osin, Dalmedigos,
  Safahi, Meirom, Belinkov, Shalev-Shwartz, Abend, Alon, Asida, Bergman,
  Glozman, Gokhman, Manevich, Ratner, Rozen, Shwartz, Zusman, and
  Shoham}]{lieber_jamba_2024}
Opher Lieber, Barak Lenz, Hofit Bata, Gal Cohen, Jhonathan Osin, Itay
  Dalmedigos, Erez Safahi, Shaked Meirom, Yonatan Belinkov, Shai
  Shalev-Shwartz, Omri Abend, Raz Alon, Tomer Asida, Amir Bergman, Roman
  Glozman, Michael Gokhman, Avashalom Manevich, Nir Ratner, Noam Rozen, and 3
  others. 2024.
\newblock Jamba: {A} {Hybrid} {Transformer}-{Mamba} {Language} {Model}.
\newblock \emph{arXiv preprint}.
\newblock ArXiv:2403.19887.
\newblock URL: \url{http://arxiv.org/abs/2403.19887}, \href
  {https://doi.org/10.48550/arXiv.2403.19887}
  {\path{doi:10.48550/arXiv.2403.19887}}.

\bibitem[{Meta({\natexlab{a}})}]{llama-3.1}
Meta. {\natexlab{a}}.
\newblock Llama 3.1.
\newblock URL:
  \url{https://www.llama.com/docs/model-cards-and-prompt-formats/llama3_1/}.

\bibitem[{Meta({\natexlab{b}})}]{llama-4-scout}
Meta. {\natexlab{b}}.
\newblock Llama 4.
\newblock URL:
  \url{https://www.llama.com/docs/model-cards-and-prompt-formats/llama4/}.

\bibitem[{Meta({\natexlab{c}})}]{llama-3.3}
Meta. {\natexlab{c}}.
\newblock meta-llama/llama-3.3-70b-instruct.
\newblock URL: \url{https://huggingface.co/meta-llama/Llama-3.3-70B-Instruct}.

\bibitem[{Meta(2024)}]{llama-3.1-release}
Meta. 2024.
\newblock Introducing {L}lama 3.1: {O}ur most capable models to date.
\newblock URL: \url{https://ai.meta.com/blog/meta-llama-3-1/}.

\bibitem[{Meta(2025)}]{llama-4-release}
Meta. 2025.
\newblock The {L}lama 4 herd: {T}he beginning of a new era of natively
  multimodal {AI} innovation.
\newblock URL: \url{https://ai.meta.com/blog/llama-4-multimodal-intelligence/}.

\bibitem[{MistralAI()}]{ministral-release}
MistralAI.
\newblock Changelog.
\newblock URL: \url{https://docs.mistral.ai/getting-started/changelog/}.

\bibitem[{MistralAI(2025{\natexlab{a}})}]{mistral}
MistralAI. 2025{\natexlab{a}}.
\newblock Medium is the new large.
\newblock URL: \url{https://mistral.ai/news/mistral-medium-3}.

\bibitem[{MistralAI(2025{\natexlab{b}})}]{mistral-medium-3.1-release}
MistralAI. 2025{\natexlab{b}}.
\newblock Mistral {M}edium 3.1.
\newblock URL: \url{https://docs.mistral.ai/models/mistral-medium-3-1-25-08}.

\bibitem[{Mousavi et~al.(2024)Mousavi, Alghisi, and
  Riccardi}]{mousavi_dyknow_2024}
Seyed~Mahed Mousavi, Simone Alghisi, and Giuseppe Riccardi. 2024.
\newblock {DyKnow}: {Dynamically} {Verifying} {Time}-{Sensitive} {Factual}
  {Knowledge} in {LLMs}.
\newblock \emph{arXiv preprint}.
\newblock ArXiv:2404.08700 [cs].
\newblock URL: \url{http://arxiv.org/abs/2404.08700}, \href
  {https://doi.org/10.48550/arXiv.2404.08700}
  {\path{doi:10.48550/arXiv.2404.08700}}.

\bibitem[{OpenAI({\natexlab{a}})}]{gpt-3.5}
OpenAI. {\natexlab{a}}.
\newblock {GPT} 3.5 {T}urbo.
\newblock URL: \url{https://platform.openai.com/docs/models/gpt-3.5-turbo}.

\bibitem[{OpenAI({\natexlab{b}})}]{gpt-4o}
OpenAI. {\natexlab{b}}.
\newblock {GPT} 4o.
\newblock URL: \url{https://platform.openai.com/docs/models/gpt-4o}.

\bibitem[{OpenAI({\natexlab{c}})}]{gpt-4o-mini}
OpenAI. {\natexlab{c}}.
\newblock {GPT}-4o mini.
\newblock URL: \url{https://platform.openai.com/docs/models/gpt-4o-mini}.

\bibitem[{OpenAI(2022)}]{gpt-3.5-release}
OpenAI. 2022.
\newblock Introducing {C}hat{GPT}.
\newblock URL: \url{https://openai.com/index/chatgpt/}.

\bibitem[{OpenAI(2024{\natexlab{a}})}]{gpt-4o-mini-release}
OpenAI. 2024{\natexlab{a}}.
\newblock {GPT}-4o mini: advancing cost-efficient intelligence.
\newblock URL:
  \url{https://openai.com/index/gpt-4o-mini-advancing-cost-efficient-intelligence/}.

\bibitem[{OpenAI(2024{\natexlab{b}})}]{gpt-4o-release}
OpenAI. 2024{\natexlab{b}}.
\newblock Hello {GPT}-4o.
\newblock URL: \url{https://openai.com/index/hello-gpt-4o/}.

\bibitem[{{OpenAI} et~al.(2025){OpenAI}, Agarwal, Ahmad, Ai, Altman, Applebaum,
  Arbus, Arora, Bai, Baker, Bao, Barak, Bennett, Bertao, Brett, Brevdo,
  Brockman, Bubeck, Chang, Chen, Chen, Cheung, Clark, Cook, Dukhan, Dvorak,
  Fives, Fomenko, Garipov, Georgiev, Glaese, Gogineni, Goucher, Gross, Guzman,
  Hallman, Hehir, Heidecke, Helyar, Hu, Huet, Huh, Jain, Johnson, Koch, Kofman,
  Kundel, Kwon, Kyrylov, Le, Leclerc, Lennon, Lessans, Lezcano-Casado, Li, Li,
  Lin, Liss, {Lily}, {Liu}, Liu, Lu, Lu, Martinovic, McCallum, McGrath,
  McKinney, McLaughlin, Mei, Mostovoy, Mu, Myles, Neitz, Nichol, Pachocki,
  Paino, Palmie, Pantuliano, Parascandolo, Park, Pathak, Paz, Peran, Pimenov,
  Pokrass, Proehl, Qiu, Raila, Raso, Ren, Richardson, Robinson, Rotsted,
  Salman, Sanjeev, Schwarzer, Sculley, Sikchi, Simon, Singhal, Song, Stuckey,
  Sun, Tillet, Toizer, Tsimpourlas, Vyas, Wallace, Wang, Wang, Watkins, Weil,
  Wendling, Whinnery, Whitney, Wong, Yang, Yang, Yasunaga, Ying, Zaremba, Zhan,
  Zhang, Zhang, Zhang, and Zhao}]{openai_gpt-oss-120b_2025}
{OpenAI}, Sandhini Agarwal, Lama Ahmad, Jason Ai, Sam Altman, Andy Applebaum,
  Edwin Arbus, Rahul~K. Arora, Yu~Bai, Bowen Baker, Haiming Bao, Boaz Barak,
  Ally Bennett, Tyler Bertao, Nivedita Brett, Eugene Brevdo, Greg Brockman,
  Sebastien Bubeck, Che Chang, and 107 others. 2025.
\newblock gpt-oss-120b \& gpt-oss-20b {Model} {Card}.
\newblock \emph{arXiv preprint}.
\newblock ArXiv:2508.10925.
\newblock URL: \url{http://arxiv.org/abs/2508.10925}, \href
  {https://doi.org/10.48550/arXiv.2508.10925}
  {\path{doi:10.48550/arXiv.2508.10925}}.

\bibitem[{Openchat({\natexlab{a}})}]{openchat-release}
Openchat. {\natexlab{a}}.
\newblock Open{C}hat: {A}dvancing {O}pen-source {L}anguage {M}odels with
  {M}ixed-{Q}uality {D}ata.
\newblock URL: \url{https://github.com/imoneoi/openchat}.

\bibitem[{Openchat({\natexlab{b}})}]{openchat}
Openchat. {\natexlab{b}}.
\newblock openchat/openchat\_3.5.
\newblock URL: \url{https://huggingface.co/openchat/openchat\_3.5}.

\bibitem[{Ovadia et~al.(2024)Ovadia, Brief, Mishaeli, and
  Elisha}]{ovadia_fine-tuning_2024}
Oded Ovadia, Menachem Brief, Moshik Mishaeli, and Oren Elisha. 2024.
\newblock Fine-{Tuning} or {Retrieval}? {Comparing} {Knowledge} {Injection} in
  {LLMs}.
\newblock In \emph{Proceedings of the 2024 {Conference} on {Empirical}
  {Methods} in {Natural} {Language} {Processing}}, pages 237--250, Miami,
  Florida, USA. Association for Computational Linguistics.
\newblock URL: \url{https://aclanthology.org/2024.emnlp-main.15}, \href
  {https://doi.org/10.18653/v1/2024.emnlp-main.15}
  {\path{doi:10.18653/v1/2024.emnlp-main.15}}.

\bibitem[{Simtheory()}]{kimi-sim}
Simtheory.
\newblock Kimi {K}2.
\newblock URL: \url{https://simtheory.ai/model-card/kimi-k2/}.

\bibitem[{Sun et~al.(2025)Sun, Aksitov, Zhmoginov, Miller, Vladymyrov,
  Rueckert, Kim, and Sandler}]{sun2025newdatapermeatesllm}
Chen Sun, Renat Aksitov, Andrey Zhmoginov, Nolan~Andrew Miller, Max Vladymyrov,
  Ulrich Rueckert, Been Kim, and Mark Sandler. 2025.
\newblock How new data permeates llm knowledge and how to dilute it.
\newblock \emph{Preprint}, arXiv:2504.09522.
\newblock URL: \url{https://arxiv.org/abs/2504.09522}, \href
  {https://arxiv.org/abs/2504.09522} {\path{arXiv:2504.09522}}.

\bibitem[{Tan et~al.(2023{\natexlab{a}})Tan, Ng, and Bing}]{tan_towards_2023}
Qingyu Tan, Hwee~Tou Ng, and Lidong Bing. 2023{\natexlab{a}}.
\newblock Towards {Benchmarking} and {Improving} the {Temporal} {Reasoning}
  {Capability} of {Large} {Language} {Models}.
\newblock In \emph{Proceedings of the 61st {Annual} {Meeting} of the
  {Association} for {Computational} {Linguistics} ({Volume} 1: {Long}
  {Papers})}, pages 14820--14835, Toronto, Canada. Association for
  Computational Linguistics.
\newblock URL: \url{https://aclanthology.org/2023.acl-long.828}, \href
  {https://doi.org/10.18653/v1/2023.acl-long.828}
  {\path{doi:10.18653/v1/2023.acl-long.828}}.

\bibitem[{Tan et~al.(2023{\natexlab{b}})Tan, Dwivedi-Yu, Li, Mathias, Saeidi,
  Yan, and Halevy}]{tan_timelineqa_2023}
Wang-Chiew Tan, Jane Dwivedi-Yu, Yuliang Li, Lambert Mathias, Marzieh Saeidi,
  Jing~Nathan Yan, and Alon Halevy. 2023{\natexlab{b}}.
\newblock {TimelineQA}: {A} {Benchmark} for {Question} {Answering} over
  {Timelines}.
\newblock In \emph{Findings of the {Association} for {Computational}
  {Linguistics}: {ACL} 2023}, pages 77--91, Toronto, Canada. Association for
  Computational Linguistics.
\newblock URL: \url{https://aclanthology.org/2023.findings-acl.6}, \href
  {https://doi.org/10.18653/v1/2023.findings-acl.6}
  {\path{doi:10.18653/v1/2023.findings-acl.6}}.

\bibitem[{xAI()}]{grok}
xAI.
\newblock Models and {P}ricing.
\newblock URL: \url{https://docs.x.ai/docs/models}.

\bibitem[{xAI(2025{\natexlab{a}})}]{grok3}
xAI. 2025{\natexlab{a}}.
\newblock Grok 3 {B}eta — {T}he {A}ge of {R}easoning {A}gents.
\newblock URL: \url{https://x.ai/news/grok-3}.

\bibitem[{xAI(2025{\natexlab{b}})}]{grok4}
xAI. 2025{\natexlab{b}}.
\newblock Grok 4 {M}odel {C}ard.
\newblock URL: \url{https://data.x.ai/2025-08-20-grok-4-model-card.pdf}.

\bibitem[{Xu et~al.(2025)Xu, Guo, He, Hu, He, Bai, Chen, Wang, Fan, Dang,
  Zhang, Wang, Chu, and Lin}]{qwen-2.5}
Jin Xu, Zhifang Guo, Jinzheng He, Hangrui Hu, Ting He, Shuai Bai, Keqin Chen,
  Jialin Wang, Yang Fan, Kai Dang, Bin Zhang, Xiong Wang, Yunfei Chu, and
  Junyang Lin. 2025.
\newblock Qwen2.5-omni technical report.
\newblock \href {https://arxiv.org/abs/arXiv:2503.20215}
  {\path{arXiv:arXiv:2503.20215}}.

\bibitem[{Xu et~al.(2020)Xu, Zhong, Yepes, and Lau}]{forget_Me_Not}
Ying Xu, Xu~Zhong, Antonio Jose~Jimeno Yepes, and Jey~Han Lau. 2020.
\newblock Forget me not: Reducing catastrophic forgetting for domain adaptation
  in reading comprehension.
\newblock In \emph{2020 International Joint Conference on Neural Networks
  (IJCNN)}, pages 1--8.
\newblock \href {https://doi.org/10.1109/IJCNN48605.2020.9206891}
  {\path{doi:10.1109/IJCNN48605.2020.9206891}}.

\bibitem[{Zhao et~al.(2025)Zhao, Awasthi, and
  Haghtalab}]{zhao2025stylefactsmappingboundaries}
Eric Zhao, Pranjal Awasthi, and Nika Haghtalab. 2025.
\newblock From style to facts: Mapping the boundaries of knowledge injection
  with finetuning.
\newblock \emph{Preprint}, arXiv:2503.05919.
\newblock URL: \url{https://arxiv.org/abs/2503.05919}, \href
  {https://arxiv.org/abs/2503.05919} {\path{arXiv:2503.05919}}.

\bibitem[{Zhu et~al.(2025{\natexlab{a}})Zhu, Chen, Gao, Zhang, Tiwari, and
  Wang}]{chenghaozhu-etal-2025-llm}
Chenghao Zhu, Nuo Chen, Yufei Gao, Yunyi Zhang, Prayag Tiwari, and Benyou Wang.
  2025{\natexlab{a}}.
\newblock Is your {LLM} outdated? a deep look at temporal generalization.
\newblock In \emph{Proceedings of the 2025 Conference of the Nations of the
  Americas Chapter of the Association for Computational Linguistics: Human
  Language Technologies (Volume 1: Long Papers)}, pages 7433--7457,
  Albuquerque, New Mexico. Association for Computational Linguistics.
\newblock URL: \url{https://aclanthology.org/2025.naacl-long.381/}, \href
  {https://doi.org/10.18653/v1/2025.naacl-long.381}
  {\path{doi:10.18653/v1/2025.naacl-long.381}}.

\bibitem[{Zhu et~al.(2025{\natexlab{b}})Zhu, Liao, Chen, Wang, Guan, Wang, and
  Wang}]{zhu-etal-2025-evolvebench}
Zhiyuan Zhu, Yusheng Liao, Zhe Chen, Yuhao Wang, Yunfeng Guan, Yanfeng Wang,
  and Yu~Wang. 2025{\natexlab{b}}.
\newblock {E}volve{B}ench: A comprehensive benchmark for assessing temporal
  awareness in {LLM}s on evolving knowledge.
\newblock In \emph{Proceedings of the 63rd Annual Meeting of the Association
  for Computational Linguistics (Volume 1: Long Papers)}, pages 16173--16188,
  Vienna, Austria. Association for Computational Linguistics.
\newblock URL: \url{https://aclanthology.org/2025.acl-long.788/}, \href
  {https://doi.org/10.18653/v1/2025.acl-long.788}
  {\path{doi:10.18653/v1/2025.acl-long.788}}.

\end{thebibliography}

\end{document}